%% file: main.tex
\documentclass[10pt,twocolumn,letterpaper]{article}

\usepackage[pagenumbers]{cvpr} % To force page numbers, e.g. for an arXiv version

\input{preamble}

\definecolor{cvprblue}{rgb}{0.21,0.49,0.74}
\usepackage[pagebackref,breaklinks,colorlinks,allcolors=cvprblue]{hyperref}

\def\confName{CVPR}
\def\confYear{2027}

\title{From Knowing to Judging Temporal Relations in\\Video LLM-as-a-Judge via Self-Generated Comparisons}

\author{%
    Jiho Choi
    \quad Inha Kang
    \quad Seojeong Park
    \quad Seungjun Yu
    \quad Hyunjung Shim\\[2pt]
    Graduate School of AI, KAIST, Republic of Korea\\
    {\tt\small \{jihochoi, rkswlsj13, seojeong.park, seungjunyu, kateshim\}@kaist.ac.kr}
}

\begin{document}
\etocdepthtag.toc{main}
\maketitle
\input{figures/fig_1_teaser}

\input{sections/0_abstract}

% Figures 2 and 3 are two-column floats: each is placed ahead of its section so that it lands at the top of the intended page.
\input{figures/fig_2_diagnosis_behavior}
\input{sections/1_introduction}

\input{sections/2_related_work}

\input{sections/3_diagnosis}
\input{figures/fig_3_method}

\input{sections/4_method}

\input{sections/5_experiments}

\input{sections/7_conclusion}
% \clearpage
\input{sections/8_llm_usage}
{
    \small
    \bibliographystyle{ieeenat_fullname}
    \bibliography{main}\par
}

% WARNING: do not forget to delete the supplementary pages from your submission
\clearpage
\setcounter{page}{1}
\maketitlesupplementary

\appendix
\setcounter{figure}{0}
\setcounter{table}{0}
\renewcommand{\thefigure}{S\arabic{figure}}
\renewcommand{\thetable}{S\arabic{table}}
\etocdepthtag.toc{appendix}
\input{sections/app_0_contents}
\clearpage
\input{sections/app_0_main}

\end{document}

%% file: preamble.tex
\usepackage{cuted} %< for strip environment

\usepackage{currfile} %< for \currfilebase macro

\usepackage{caption} %< for "\captionof" commands (teaser)
\usepackage{needspace}
\usepackage{colortbl}
\usepackage{silence}
\usepackage{etoc}    %< contents list of the supplementary material
\usepackage{tikz}    %< prompt boxes of the supplementary material
\usepackage{paper-notation}    %< math notation shared by the paper and the supplementary material

\definecolor{plotblue}{HTML}{276995}
\definecolor{deltaup}{HTML}{1B7A3E}       %< gain over the base model
\definecolor{deltadown}{HTML}{B8322A}     %< drop from the base model
\definecolor{deltazero}{RGB}{200,200,200} %< no change
\definecolor{oursrow}{HTML}{EAF3FA}       %< row of our method
\definecolor{bandrow}{HTML}{EEF3F8}       %< row that names a group

\newcommand{\dplus}[1]{{\tiny\textcolor{deltaup}{(+#1)}}}
\newcommand{\dminus}[1]{{\tiny\textcolor{deltadown}{($-$#1)}}}
\newcommand{\dns}[1]{{\tiny\textcolor{deltazero}{(#1)}}}
\newcommand{\vdc}[2]{\vtop{\halign{\hfil##\hfil\cr\strut#1\cr\noalign{\vskip-3.4pt}\strut#2\cr}}}
\newcommand{\dcolp}[1]{{\scriptsize\textcolor{deltaup}{(+#1)}}}
\newcommand{\dcolm}[1]{{\scriptsize\textcolor{deltadown}{($-$#1)}}}
\newcommand{\dcolz}{{\scriptsize\textcolor{deltazero}{(0.0)}}}

\newcommand{\method}{\mbox{\upshape\textsc{TemporalJudge}}\xspace}
\newcommand{\claimex}[1]{\textit{``#1''}}

%% file: figures/fig_1_teaser.tex
\begin{strip}
\centering
\includegraphics[width=\textwidth]{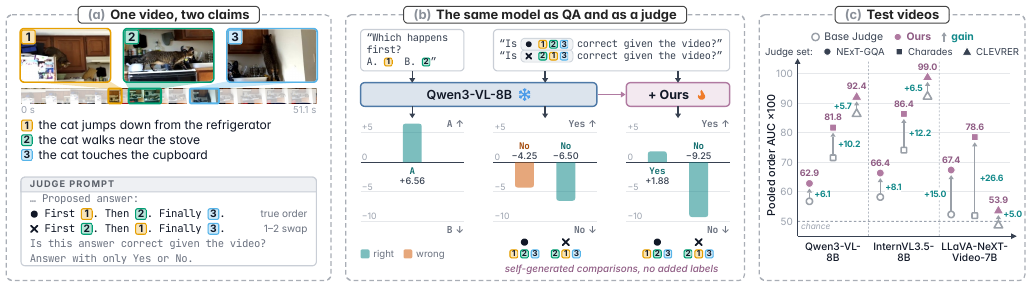}
\captionsetup{hypcap=false}

\captionof{figure}{
    \textbf{Answering a temporal order question does not ensure correct judgment.}
    (a) A video and two claims describing the same events in different orders.
    (b) Qwen3-VL-8B identifies the correct order in QA but rejects the true claim in standalone judgment.
    Our method (\method) accepts the true claim and rejects the reordered negative claim.
    Bars show answer-token logit differences, averaged over both option orders for QA and computed as \tokYes{} minus \tokNo{} for judgment.
    (c) \method improves pooled order AUC across video LLM backbones and judge sets.
}

\captionsetup{hypcap=true}
\label{fig:Answering_an_order_question_does_not_ensure_a_correct_judgment}%
\vspace{-1.0em}
\end{strip}

%% file: sections/0_abstract.tex
\begin{abstract}
Video large language models (video LLMs) are increasingly used as automated judges for scalable assessment of video-grounded responses.
Reliable judgment requires verifying both event content and temporal relations.
However, even when video LLMs identify the correct event order in comparative question answering, they can still fail when judging standalone claims.
We investigate this discrepancy through representation probing and score decomposition.
Although temporal information is partially accessible in hidden states, standalone judgment scores primarily reflect event-content consistency.
Within a video, order preferences remain weak.
Across videos, variation in shared score offsets further obscures temporal differences.
To address this limitation, we introduce \method, a framework that derives temporal supervision from self-generated, content-preserving comparisons.
Using confidence-filtered descriptions of chronological segments, we construct true claims and reordered negatives that share identical event descriptions.
Comparing these claims within each video mitigates the influence of shared score offsets and directly supervises temporal order.
A complementary content anchor helps preserve sensitivity to errors in event content.
Training requires no human-annotated claims or temporal labels, while inference evaluates each claim independently.
Evaluated across multiple video LLMs and benchmarks, \method consistently improves the judgment of temporal relations.
These gains also transfer to coarse temporal localization and connective-based temporal claims.
\end{abstract}

% TemporalJudge

% Keywords
% Video LLM-as-a-Judge,Temporal Reasoning,Temporal Consistency,Self-Generated Comparisons
% Self-Improving Learning

% 2026.11
% Reliable video judges must verify event content and temporal relations.
% We find that video LLMs identify event order more reliably than they distinguish correct claims from reordered counterparts under single-claim inference.
% Our analysis finds stronger discrimination of event content than temporal order.
% We identify weak within-video order preferences and shared score offsets that can obscure order discrimination across videos.
% We propose \method, a framework that learns temporal judgment from self-generated, content-preserving comparisons within each video without human-annotated training claims.
% The training objective directly supervises order preferences and includes a content anchor to help preserve content discrimination.
% At inference, the trained judge scores each video-claim pair independently.
% Across three backbones, \method improves mean pooled order AUC by 7.3 to 15.5 points over the base models.
% The gains extend to coarse temporal localization and external benchmarks.

%% file: figures/fig_2_diagnosis_behavior.tex
\begin{figure*}[t]
\centering
\includegraphics[width=\textwidth]{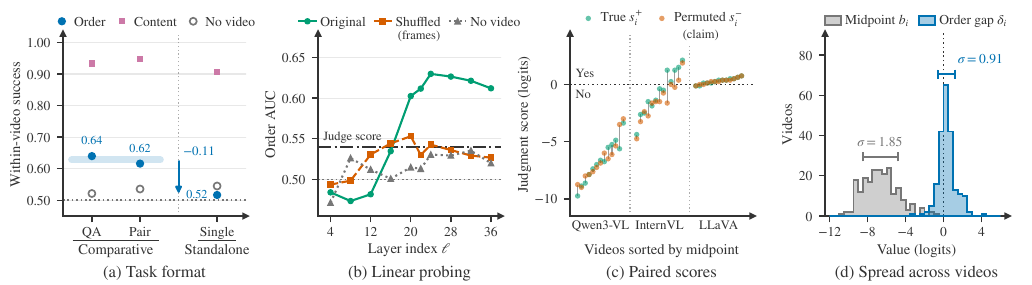}
\caption{
    \textbf{Weak temporal order sensitivity in judgment.}
    Diagnostics on the NExT-GQA~\cite{xiao2024nextgqa} validation set.
    (a) Temporal discrimination is stronger in comparative (QA, Pair) than standalone (Single) formats, while content discrimination remains high.
    (b) Linear probes on frozen hidden states distinguish temporal order more reliably than judgment scores; shuffling or removing frames reduces probe performance.
    (c) Chronological and reordered claims exhibit similar score trends across videos.
    (d) The shared midpoint $b_i$ varies more across videos than the order gap $\delta_i$ (\Cref{eq:decomposition}); $\sigma$ denotes standard deviation.
    Additional experimental details are provided in \Cref{app:Extended_Diagnostics}.
}
\label{fig:Weak_Temporal_Order_Sensitivity_in_Judgment}
\end{figure*}

%% file: sections/1_introduction.tex
\section{Introduction}
\label{sec:Introduction}

% ----------------------------------------------------------------
% (1.1 - 1.4) 동기 + 한계 + 발견 + 분석
% ----------------------------------------------------------------

% 1.1. 연구 배경 및 필요성 (Motivation)
Recent advances in video large language models (video LLMs) have significantly improved their capabilities in video understanding and reasoning~\cite{zhang2024llavanextvideo,bai2025qwen3,wang2025internvl35}.
Beyond video question answering (Video QA), video LLMs are increasingly used as judges to evaluate video-conditioned responses and provide reward signals~\cite{liu2025your,wang2025unifiedreward,waheed2025videojudge}.
Reliable video judgment requires verifying whether a claim accurately describes the video, including its event content and temporal relations.
For instance, \Cref{fig:Answering_an_order_question_does_not_ensure_a_correct_judgment}\textcolor{cvprblue}{(a)} shows a cat jumping down from a refrigerator before walking near a stove.
A claim stating \claimex{First the cat walks near the stove. Then the cat jumps down from the refrigerator.} describes the correct events but reverses their temporal order.
Such inconsistencies are also relevant to evaluating video generation and world models~\cite{zheng2025vbench2,nvidia2025cosmos}, where individually plausible events can form an incorrect sequence.
Thus, reliable video judgment requires distinguishing correct event content from correct temporal relations.

% 1.2. 기존 한계 (Limitations of Existing Approaches)
Despite advances in temporal video understanding and multimodal evaluation, reliably assessing temporal consistency in individual claims remains challenging.
Temporal reasoning is commonly evaluated through video question answering benchmarks~\cite{xiao2021nextqa,xiao2024nextgqa,liu2024tempcompass}, but correctly answering temporal questions does not necessarily translate into reliable judgment of corresponding claims, as illustrated in \Cref{fig:Answering_an_order_question_does_not_ensure_a_correct_judgment}\textcolor{cvprblue}{(b)}.
Meanwhile, general-purpose video judges are trained to assess overall response correctness or quality~\cite{xiong2025llavacritic,wang2025unifiedreward,waheed2025videojudge,lin2025selfjudges}, without explicitly distinguishing temporal inconsistencies from errors in event content.
Specialized approaches such as TimeJudge~\cite{hu2025timejudge} assess temporal consistency through auxiliary verification questions and answer aggregation, rather than directly evaluating individual claims.
VELOCITI~\cite{saravanan2025velociti} further shows that strict video-caption entailment is more challenging than relative caption ranking.
However, why standalone temporal judgment is unreliable and how to improve it remain underexplored.

% 1.3. Key Observation
To investigate this limitation, we compare video question answering and standalone judgment on the same temporal-order fact (\Cref{sec:Discrepancy_Between_Temporal_Understanding_and_Judgment}).
We find that a video LLM identifies the order of events more reliably than it distinguishes correct claims from their reordered counterparts when scored independently.
\Cref{fig:Answering_an_order_question_does_not_ensure_a_correct_judgment}\textcolor{cvprblue}{(b)} illustrates a related failure, where the model identifies the correct order but rejects a claim describing that sequence.
Furthermore, linear probes on frozen hidden states~\cite{alain2016probes,hewitt2019probes} distinguish temporal order more reliably than the model's judgment scores (\Cref{fig:Weak_Temporal_Order_Sensitivity_in_Judgment}\textcolor{cvprblue}{(b)}).
These findings suggest that temporal information is partially accessible in the model but not reliably reflected in standalone judgment scores.

% 1.4. Analysis and Insight
To understand this discrepancy, we analyze how temporal relations are reflected in the judgment scores of correct claims and their reordered counterparts (\Cref{sec:Order_Sensitivity_and_Score_Offsets}).
Our analysis suggests that judgment scores reflect event-content consistency more reliably than temporal order.
Within a video, the claims share the same event content, yet the judge often shows weak or incorrect order preferences.
Across videos, a shared score offset further obscures these differences (\Cref{fig:Weak_Temporal_Order_Sensitivity_in_Judgment}\textcolor{cvprblue}{(c,d)}).
These findings motivate learning from within-video comparisons, which cancel shared offsets and directly supervise temporal order discrimination.

% ----------------------------------------------------------------
% (1.5 - 1.7) 제안 방법 + 주요 실험 결과 + 기여
% ----------------------------------------------------------------

% 1.5. Proposed Method
Building on this insight, we propose \method, a framework that learns temporal judgment from self-generated within-video comparisons (\Cref{fig:Self_improving_temporal_judgment}).
A frozen base model first describes video segments sampled in chronological order, and we select records based on segment-description correspondence.
We then construct contrasting claims by preserving event content while varying their temporal order.
The judge learns to distinguish temporally consistent claims from their reordered counterparts through within-video comparisons, which cancel the shared score offset and directly supervise the order gap.
To help preserve content discrimination, we additionally introduce a content anchor that constrains the scores of claims containing incorrect events.
The framework requires no human-written event descriptions or temporal annotations for training.
At inference, the trained judge evaluates each video-claim pair independently, without comparing alternative claims.

% 1.6. Experimental Results
Experiments on three video judgment datasets demonstrate that \method improves temporal discrimination while preserving strong content discrimination.
The improvements are consistent across multiple video LLM backbones (\Cref{fig:Answering_an_order_question_does_not_ensure_a_correct_judgment}(c)).
Further analysis shows that training strengthens temporal order preferences relative to cross-video score variation, mitigating the limitations identified in our diagnosis.
Beyond event order, the improvements extend to other temporal judgments, including coarse temporal localization, and generalize to external benchmarks such as TempCompass~\cite{liu2024tempcompass}, VELOCITI~\cite{saravanan2025velociti}, and Vinoground~\cite{zhang2024vinoground}.

% 1.7. Contributions
Our main contributions are as follows:
\begin{itemize}
    \item We identify and analyze the discrepancy between temporal
    understanding and standalone judgment.
    Our analysis distinguishes weak within-video order preferences from
    cross-video variation in shared score offsets.

    \item We introduce \method, learning temporal judgment from 
    self-generated, content-preserving comparisons within each video without 
    human-annotated training claims, retaining single-claim inference.

    \item We demonstrate consistent gains in temporal order discrimination
    across video LLMs and benchmarks, with transfer to coarse temporal
    localization.
\end{itemize}

%% file: sections/2_related_work.tex
\section{Related Work}
\label{sec:Related_Work}

\paragraph{Video LLM-as-a-Judge.}
LLM-as-a-Judge has been increasingly adopted for multimodal evaluation, including video-based evaluation, with growing attention to judgment reliability~\cite{zheng2023judging,chen2024mllmjudge,pu2025judgeanything,liu2025your,wang2025trustjudge}.
Prior work has explored supervised fine-tuning, reward modeling, and synthetic supervision to improve multimodal evaluation~\cite{xiong2025llavacritic,lee2024prometheusvision,wang2025unifiedreward,ko2025flexjudge}.
VideoJudge~\cite{waheed2025videojudge} employs generator-evaluator bootstrapping to train video-conditioned evaluators.
Self-improving VLM judges~\cite{lin2025selfjudges} iteratively learn from synthesized supervision without human preference labels, building on self-taught and self-rewarding evaluation~\cite{wang2024selftaught,yuan2024selfrewarding}.
Recent work has also investigated perceptual judgment biases and temporal consistency in multimodal evaluation~\cite{ianaro2026ordermatters,li2026isovrbench,park2026perceptionjudge}.
In particular, TimeJudge~\cite{hu2025timejudge} improves temporal consistency evaluation through auxiliary verification questions and answer aggregation.
In contrast, our approach directly learns temporal judgment from self-generated, content-preserving comparisons while retaining single-claim inference.

\paragraph{Temporal Reasoning and Probing.}
Temporal understanding in video LLMs has been evaluated through question answering, temporal grounding, and compositional reasoning benchmarks~\cite{xiao2021nextqa,xiao2024nextgqa,liu2024tempcompass,li2026timeblind}.
Other benchmarks evaluate video-text consistency through controlled caption contrasts and entailment tasks~\cite{li2024vitatecs,zhang2024vinoground,saravanan2025velociti}.
Prior studies have also identified limitations in compositional understanding and answer biases~\cite{yuksekgonul2022and,li2024naturalbench}, while investigating how temporal information is encoded in video LLMs~\cite{han2026tracing}.
Probing methods further reveal information encoded in model representations that may not be reflected in their outputs~\cite{alain2016probes,hewitt2019probes,burns2023latent}.
Building on these studies, we investigate whether temporal information is reflected in standalone judgment scores through controlled task comparisons and probing of frozen model representations.

\paragraph{Learning from Generated Video Data.}
Temporal perturbations have been widely used to improve video understanding through contrastive learning, entailment training on contrast captions, preference optimization, and reward-based training~\cite{bagad2023test,bansal2024videocon,li2026temple,kulkarni2025videopasta,li2025tpo,feng2025videor1,xue2025arrowoftime}.
Recent approaches further explore self-supervised and self-evolving video learning using temporal pretext tasks, perturbation sensitivity, and segment-level supervision~\cite{he2026videossr,huang2026evovid,zhang2026videozero}.
However, improvements on temporal benchmarks do not always imply sensitivity to the actual order of video frames~\cite{sadhu2027verifiable}.
Self-generated contrasts have also been explored for multimodal reward learning without human preference labels~\cite{li2026unsupervisedmrm}.
Unlike approaches primarily targeting video understanding or general reward learning, our framework targets standalone temporal judgment.
Our framework preserves event content while varying temporal order, selects video-grounded examples, and constrains content-negative scores to maintain content discrimination.

%% file: sections/3_diagnosis.tex
\section{Diagnosing Temporal Judgment}
\label{sec:Diagnosing_Temporal_Judgment}

% ----------------------------------------------------------------
% 3.1 Task and Controlled Evaluation
% ----------------------------------------------------------------

\subsection{Task and Controlled Evaluation}
\label{sec:Task_and_Controlled_Evaluation}

Given a video $v$ and a claim $c$, a video judge assesses whether the claim is supported by the video.
We quantify this judgment using the model's output logits.
Let $\pi_\theta$ be a video LLM with parameters $\theta$, and $z_\theta(t \mid v,c;q)$ denote the logit for token $t$ at the first response position under prompt $q$.
Following entailment verification~\cite{sanyal2024machines,saravanan2025velociti}, we define the judgment score as
\begin{equation}
    s_\theta(v,c;q)=z_\theta(\tokYes \mid v,c;q) - z_\theta(\tokNo \mid v,c;q),
    \label{eq:score}
\end{equation}
and write $s_\theta(v,c)$ for the fixed judgment prompt $q_{\mathtt{judge}}$.
A larger score indicates stronger support for the claim.
We evaluate judgment by how reliably these scores distinguish correct from incorrect claims.

To isolate temporal judgment from event-content recognition, we construct controlled claims for each video $v_i$.
A positive claim $c_i^{+}$ describes events in their chronological order, while an \emph{order negative} $c_i^{-}$ rearranges the same events without changing their content.
A \emph{content negative} $c_{i,\mathtt{content}}^{-}$ instead replaces an event with one from another video.
We denote the scores of the positive claim and its order negative by $s_i^{+}$ and $s_i^{-}$, respectively.
We assess judgment using pooled AUC across independently scored claims and within-video comparisons between claims from the same video (\Cref{app:Metrics_and_Evaluation_Protocol}).

% ----------------------------------------------------------------
% 3.2 Temporal Understanding vs. Judgment
% ----------------------------------------------------------------

\subsection{Temporal Understanding vs. Judgment}
\label{sec:Discrepancy_Between_Temporal_Understanding_and_Judgment}

\paragraph{Temporal discrimination weakens in standalone judgment.}
We compare the same temporal fact, the order of two events, across three task formats using the same videos and model (\Cref{fig:Weak_Temporal_Order_Sensitivity_in_Judgment}\textcolor{cvprblue}{(a)}).
The comparative formats ask which event occurs first (QA) or which of two claims describing opposite orders is correct (Pair).
In contrast, Single scores each claim independently using a Yes/No judgment prompt, with temporal discrimination assessed by comparing their scores.
The model performs relatively better in QA (64\%) and Pair (62\%) than in Single (52\%).
Even when QA succeeds, Single can fail to distinguish the corresponding claims (\Cref{app:Extended_Diagnostics}).
Meanwhile, content discrimination, which distinguishes correct events from incorrect ones, exceeds 90\% across all formats.
These results show that temporal order is substantially harder to discriminate than event content, with further degradation in standalone judgment compared with comparative formats.
While prior work has highlighted generator-validator inconsistency~\cite{li2024benchmarking} and limitations of strict video-caption entailment~\cite{saravanan2025velociti}, our analysis isolates the discrepancy between identifying temporal order and independently judging claims about that order.

\paragraph{Hidden representations contain accessible temporal information.}
To investigate whether this discrepancy reflects a lack of temporal information, we train linear probes~\cite{alain2016probes} on frozen hidden states and evaluate them on held-out videos.
Using separate three-event claims from the same validation set, the probes distinguish correct temporal order from its permutations more reliably than the model's judgment scores (\Cref{fig:Weak_Temporal_Order_Sensitivity_in_Judgment}(b)).
Probe performance declines substantially when video frames are shuffled or removed, suggesting that the learned distinction relies on visual temporal information.
These findings suggest that temporal information is partially accessible in the model's representations but not reliably reflected in its judgment scores~\cite{burns2023latent}.
However, probing alone cannot establish whether the judge uses this information.

% ----------------------------------------------------------------
% 3.3 Order Sensitivity and Score Offsets
% ----------------------------------------------------------------

\subsection{Order Sensitivity and Score Offsets}
\label{sec:Order_Sensitivity_and_Score_Offsets}

The stronger discrimination of event content than temporal order suggests that judgment scores are more sensitive to event-content consistency.
To understand how this imbalance affects temporal judgment, we examine score differences within videos and score variation across videos.
For each video, we decompose the scores of a positive claim and its order negative into a shared midpoint $b_i$ and a signed order gap $\delta_i$:
\begin{equation}
    b_i = \frac{s_i^{+}+s_i^{-}}{2}, \quad
    \delta_i = s_i^{+}-s_i^{-}, \quad
    s_i^{\pm} = b_i\pm\frac{\delta_i}{2}.
\label{eq:decomposition}
\end{equation}
The gap $\delta_i$ measures the judge's preference for the correct temporal order, while $b_i$ represents the shared score level.

\paragraph{Within-video order sensitivity.}
When comparing claims from the same video, the shared midpoint cancels, so temporal discrimination depends only on the order gap $\delta_i$.
The weak standalone judgment observed in \Cref{sec:Discrepancy_Between_Temporal_Understanding_and_Judgment} therefore reflects weak or incorrect preferences between temporal orders.

\paragraph{Cross-video score variation.}
When comparing claims across videos, differences in their shared midpoints can further obscure temporal order.
As shown in \Cref{fig:Weak_Temporal_Order_Sensitivity_in_Judgment}(c), correct claims and their reorderings exhibit similar score trends despite substantial variation across videos.
The midpoint varies approximately twice as much as the order gap across videos (\Cref{fig:Weak_Temporal_Order_Sensitivity_in_Judgment}(d)).
Consequently, a reordered claim from one video may receive a higher score than a correct claim from another.
Removing the shared midpoint raises pooled order AUC from 0.54 to 0.69 (\Cref{app:Extended_Diagnostics}), suggesting that variation in shared score levels obscures temporal order differences across videos.

These findings motivate learning from within-video comparisons, where shared offsets cancel and the order gap can be directly supervised.
Unlike removing offsets at inference, this approach retains independent evaluation of each video-claim pair (\Cref{sec:Self_Improving_Temporal_Judgment}).
We further examine its effect on cross-video score comparability in \Cref{sec:Analysis_of_Judgment_Improvements}.

%% file: figures/fig_3_method.tex
\begin{figure*}[t]
    \centering
    \includegraphics[width=\textwidth]{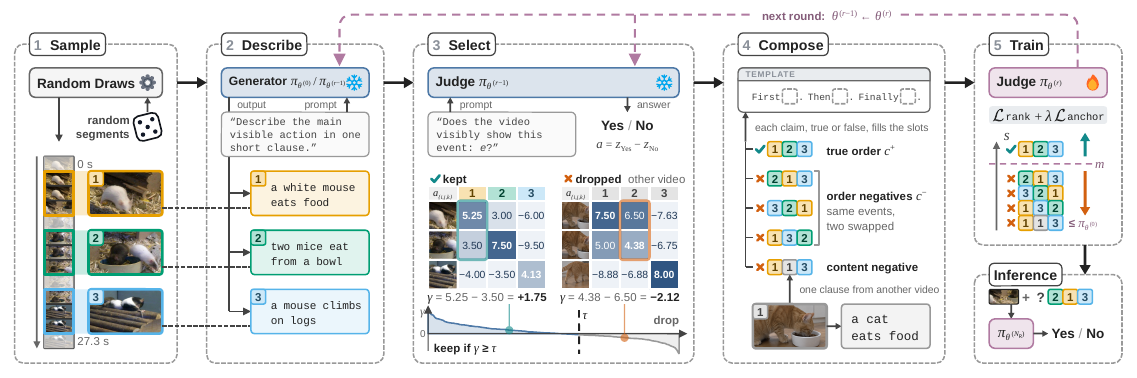}
    \vspace{-2em}
    \caption{
        \textbf{Overview of \method.}
        The framework learns temporal discrimination from within-video contrasts constructed using segment-grounded event descriptions.
        The correspondence matrix illustrates confidence-based record selection.
        The purple dashed arrow passes the trained judge to the next round as the selector and, in place of the base model, as a possible generator.
        Inference uses the judge of the last round, $\pi_{\theta^{(N_{\mathtt{R}})}}$.
    }
    \vspace{-1em}
\label{fig:Self_improving_temporal_judgment}
\end{figure*}

%% file: sections/4_method.tex
\section{Self-Generated Temporal Comparisons}
\label{sec:Self_Improving_Temporal_Judgment}
% ----------------------------------------------------------------
% 4.0 Overview
% ----------------------------------------------------------------

We propose \method, a framework that learns temporal judgment from self-generated within-video comparisons (\Cref{fig:Self_improving_temporal_judgment}).
We sample video segments in chronological order, describe them with a generator, and use the judge from the previous round $\pi_{\theta^{(r-1)}}$ to select records based on their segment-description correspondence.
The selected descriptions are composed into chronological claims and order negatives that preserve event content while changing temporal order.
Since the order of sampled segments is known, this process provides temporal supervision without human-written event descriptions or temporal annotations for the training claims.

We train the judge to distinguish correct temporal orders from their reordered counterparts through within-video comparisons, while a content anchor helps preserve content discrimination.
At inference, the trained judge evaluates each video-claim pair independently using \Cref{eq:score}.

% ----------------------------------------------------------------
% 4.1 Sampling and Describing Segments
% ----------------------------------------------------------------
\subsection{Sampling and Describing Segments}
\label{sec:Sampling_and_Describing_Segments}

For each source video $v\in\mathcal{D}_{\mathtt{source}}$, we sample $N_{\mathtt{E}}=3$ temporal segments chronologically (step~1 in \Cref{fig:Self_improving_temporal_judgment}).
In round $r$, a generator, either the base model $\pi_{\theta^{(0)}}$ or the previous judge $\pi_{\theta^{(r-1)}}$ (\Cref{sec:Iterating_the_Self_Improving_Loop}), describes the event in each sampled segment (step~2).
We repeat this sampling and description process $N_{\mathtt{G}}=3$ times per video, producing records indexed by $i$.
Each record contains the source video $v_i=v$, sampled segments $u_{\{i,1\}},\ldots,u_{\{i,N_{\mathtt{E}}\}}$, and their descriptions $e_{\{i,1\}},\ldots,e_{\{i,N_{\mathtt{E}}\}}$.
Since generated descriptions may be inaccurate or ambiguous, we select records by their segment-description correspondence, as described next.

% ----------------------------------------------------------------
% 4.2 Selecting Segment-Grounded Records
% ----------------------------------------------------------------
\subsection{Selecting Segment-Grounded Records}
\label{sec:Selecting_Segment_Grounded_Records}

To reduce ambiguity in generated event descriptions, we evaluate their correspondence to the sampled video segments (step~3 in \Cref{fig:Self_improving_temporal_judgment}).
In round $r$, the previous judge $\pi_{\theta^{(r-1)}}$, frozen during selection, computes a presence score for every segment-description pair using a segment-level prompt $q_{\mathtt{presence}}$:
\begin{equation}
    a_{\{i,j,k\}} = s_{\theta^{(r-1)}}(u_{\{i,j\}},e_{\{i,k\}};q_{\mathtt{presence}}).
\label{eq:presence}
\end{equation}
The diagonal score $a_{\{i,j,j\}}$ measures how well segment $j$ matches its own description.
We define the confidence of each record as the minimum margin between a diagonal score and its strongest competitor in the same row or column:
\begin{equation}
    \gamma_i
    =
    \min_{\substack{1\le j,k\le N_{\mathtt{E}}\\j\ne k}}
    \left[
        a_{\{i,j,j\}}
        -
        \max\left(a_{\{i,j,k\}},a_{\{i,k,j\}}\right)
    \right].
\label{eq:confidence}
\end{equation}
As illustrated in \Cref{fig:Self_improving_temporal_judgment}, records with ambiguous segment-description correspondence receive low confidence.
For each source dataset $d$ and sampling draw, we retain records with $\gamma_i\ge\tau_d$, where $\tau_d$ is the $(1-\rho)$ empirical quantile of confidence scores and $\rho=0.6$ (\Cref{app:Data_Construction_and_Semantic_Validity}).
The threshold is computed separately for each round, source, and draw.
This selection reduces ambiguity in segment-description correspondence but does not guarantee semantic correctness (\Cref{app:Data_Construction_and_Semantic_Validity}).

% ----------------------------------------------------------------
% 4.3 Constructing Contrastive Claims
% ----------------------------------------------------------------
\subsection{Constructing Contrastive Claims}
\label{sec:Composing_Claims_and_Training_the_Judge}

For each selected record, we compose a chronological claim $c_i^{+}$ using a template (step~4 in \Cref{fig:Self_improving_temporal_judgment}).
We construct order negatives by swapping pairs of event descriptions, denoted by $\Swap(c;j,k)$:
\begin{equation}
\begin{aligned}
    c_i^{+} &= \Template(e_{\{i,1\}},\ldots,e_{\{i,N_{\mathtt{E}}\}}),\\
    \mathcal{N}_i &= \{\Swap(c_i^{+};j,k) : 1\le j<k\le N_{\mathtt{E}}\}.
\end{aligned}
\label{eq:construction}
\end{equation}
Each order negative $c_i^{-}\in\mathcal{N}_i$ preserves the event descriptions while changing their temporal order.
Training uses all $\binom{N_{\mathtt{E}}}{2}=3$ order negatives.
We also construct a candidate content negative $c_{i,\mathtt{content}}^{-}$ by replacing the middle description with one generated from another video.
The previous-round judge screens the substituted event using segment-level presence scores, and the content negative is retained only if it passes this check (\Cref{app:Data_Construction_and_Semantic_Validity}).

% ----------------------------------------------------------------
% 4.4 Learning from Within-Video Comparisons
% ----------------------------------------------------------------
\subsection{Learning from Within-Video Comparisons}
\label{sec:Learning_from_Within_Video_Comparisons}

We train the judge to score chronological claims higher than their order negatives (step~5).
For each order negative $c\in\mathcal{N}_i$, we define the score gap $\delta_i(c)=s_\theta(v_i,c_i^{+})-s_\theta(v_i,c)$, which cancels the shared offset in \Cref{eq:decomposition}.
We encourage each gap to exceed a margin $m>0$ using $\softplus(x)=\log(1+\exp(x))$:
\begin{equation}
    \mathcal{L}_{\mathtt{rank},i}
    =
    \frac{1}{|\mathcal{N}_i|}
    \sum_{c\in\mathcal{N}_i}
    \softplus\!\left(m-\delta_i(c)\right).
\label{eq:ranking}
\end{equation}

To help preserve content discrimination during order-focused training, we introduce a content anchor that penalizes increases in content-negative scores relative to the frozen base judge:
\begin{equation}
    \mathcal{L}_{\mathtt{anchor},i}
    =
    \max\!\big(
        0,\;
        s_\theta(v_i,c_{i,\mathtt{content}}^{-})
        -s_{\theta^{(0)}}(v_i,c_{i,\mathtt{content}}^{-})
    \big),
\label{eq:anchor}
\end{equation}
where $s_{\theta^{(0)}}$ is computed without gradients.
The anchor term is zero when no content negative is available.

The overall objective combines the ranking and content-anchor losses with auxiliary supervision:
\begin{equation}
    \mathcal{L}_i
    = \mathcal{L}_{\mathtt{rank},i}
    + \lambda_{\mathtt{anchor}}\mathcal{L}_{\mathtt{anchor},i}
    + \mathcal{L}_{\mathtt{aux},i}.
\label{eq:loss}
\end{equation}
Here, $\mathcal{L}_{\mathtt{aux},i}$ combines calibration, content, and no-video losses to supervise absolute claim scores, rejection of content negatives, and uniform preferences without video, respectively.
The objective is averaged over selected records, with loss weights and normalization detailed in \Cref{app:Implementation_Details}.

% ----------------------------------------------------------------
% 4.5 Iterating the Self-Improving Loop
% ----------------------------------------------------------------
\subsection{Iterating the Self-Improving Loop}
\label{sec:Iterating_the_Self_Improving_Loop}

Starting from the base judge $\pi_{\theta^{(0)}}$, each round uses the previous judge to select the training records of the next judge.
Let $\Describe$, $\Select$, and $\Train$ denote segment description, record selection, and claim construction and judge training, respectively.
For $r=1,\ldots,N_{\mathtt{R}}$, the procedure is
\begin{equation}
\begin{aligned}
    \mathcal{D}_{\mathtt{generated}}^{(r)}
    &= \Describe\big(\mathcal{D}_{\mathtt{source}},\theta^{(r-1)}\big),\\
    \mathcal{D}_{\mathtt{train}}^{(r)}
    &= \Select\big(\mathcal{D}_{\mathtt{generated}}^{(r)},\theta^{(r-1)}\big),\\
    \theta^{(r)}
    &= \Train\big(\mathcal{D}_{\mathtt{train}}^{(r)};\theta^{(0)}\big).
\end{aligned}
\label{eq:loop}
\end{equation}
The sampled segments are fixed across rounds, and the generator in $\Describe$ is either the previous judge, as written, or the base model, which reuses the first-round descriptions.
Each round initializes the student from $\theta^{(0)}$, also the frozen content-anchor reference.
Both generator choices coincide in the first round, used for the main results; \Cref{app:Additional_Ablations} compares them in later rounds.

\input{tables/tab_1_judge_sets}

\input{tables/tab_2_main_results}

%% file: tables/tab_1_judge_sets.tex
\begin{table}[t]
\centering
\footnotesize
\setlength{\tabcolsep}{2.4pt}
\caption{
    \textbf{Judge sets.}
    NG-J is used only for evaluation.
    Video lengths are test-set medians, and claim counts correspond to the test.
    Temporal claim types (Loc./Dur./Freq.) are in \Cref{app:Judge_Evaluation_Sets}.
}
\label{tab:Judge_sets}
\resizebox{\columnwidth}{!}{%
    \begin{tabular}{llrrrrrrrrr}
        \toprule
         & & \multicolumn{4}{c}{Videos} & \multicolumn{5}{c}{Claims}\\
        \cmidrule(lr){3-6}\cmidrule(lr){7-11}
         Set & Events from & Train & Val & Test & Length\,(s) & Order & Content & Loc. & Dur. & Freq.\\
        \midrule
        NG-J & QA pairs & 0 & 211 & 305 & 41.1 & 1,830 & 610 & 192 & 128 & --\\
        Ch-J & action classes & 700 & 125 & 330 & 31.2 & 1,980 & 598 & 592 & 356 & --\\
        Cl-J & collisions & 793 & 198 & 331 & 5.12 & 1,986 & 662 & 956 & 1,316 & 1,980\\
        \bottomrule
    \end{tabular}}
    \vspace{-1em}
\end{table}

%% file: tables/tab_2_main_results.tex
\begin{table*}[t]
    \centering
    \footnotesize
    \setlength{\tabcolsep}{2.3pt}
    \caption{
        \textbf{Temporal judgment and content discrimination.}
        Pooled test AUC ($\times 100$) for order (Ord.), localization (Loc.), duration (Dur.), frequency (Freq.), and content.
        $\dagger$ denotes pairwise accuracy (\%) from joint two-claim comparison, shown for reference.
        Parentheses give changes from Base; \textbf{bold} and \underline{underline} mark the best and second-best AUCs per group.
        Protocols are in \Cref{app:Metrics_and_Evaluation_Protocol,app:Baseline_Adaptations_and_Inference_Cost}.
    }
    \label{tab:Temporal_judgment_and_content_discrimination}
    \resizebox{\textwidth}{!}{%
    \begin{tabular}{lccccccccccccccccc}
        \toprule
        & \multicolumn{4}{c}{NExT-GQA-Judge~(NG-J)} & \multicolumn{4}{c}{Charades-Judge~(Ch-J)} & \multicolumn{5}{c}{CLEVRER-Judge~(Cl-J)} & \multicolumn{4}{c}{\textit{Average}}\\
        \cmidrule(lr){2-5}\cmidrule(lr){6-9}\cmidrule(lr){10-14}\cmidrule(lr){15-18}
        Method & Ord. & Loc. & Dur. & Content & Ord. & Loc. & Dur. & Content & Ord. & Loc. & Dur. & Freq. & Content & Ord. & Loc. & Dur. & Content\\
        \midrule
        LLaVA-Critic-7B~\cite{xiong2025llavacritic} & \textbf{54.5} & \underline{57.3} & \textbf{58.1} & \underline{70.2} & \textbf{61.0} & \underline{64.0} & \textbf{60.1} & \underline{63.7} & 55.7 & 54.2 & 50.5 & 52.5 & 70.5 & 57.1 & \underline{58.5} & \textbf{56.3} & 68.1\\
        UnifiedReward-7B~\cite{wang2025unifiedreward} & 51.4 & 53.6 & 57.0 & 56.9 & 54.1 & 57.6 & 50.8 & 58.2 & \underline{66.7} & \underline{62.9} & \textbf{53.3} & \textbf{53.8} & \underline{96.2} & \underline{57.4} & 58.0 & 53.7 & \underline{70.4}\\
        UnifiedReward-7B + TrustJudge~\cite{wang2025trustjudge} & \underline{52.7} & \textbf{57.7} & \underline{57.4} & \textbf{75.7} & \underline{58.1} & \textbf{67.0} & \underline{56.3} & \textbf{72.1} & \textbf{74.3} & \textbf{69.6} & \underline{52.5} & \underline{53.4} & \textbf{99.5} & \textbf{61.7} & \textbf{64.8} & \underline{55.4} & \textbf{82.4}\\
        VideoJudge-7B~\cite{waheed2025videojudge} & 50.5 & 50.1 & 56.2 & 56.2 & 50.4 & 56.0 & 54.7 & 51.2 & 50.0 & 52.1 & 47.8 & 48.6 & 50.0 & 50.3 & 52.7 & 52.9 & 52.5\\
        VideoJudge-7B + TrustJudge~\cite{wang2025trustjudge} & 52.0 & 54.1 & 52.3 & 62.3 & 52.1 & 55.6 & 54.5 & 54.6 & 54.1 & 50.3 & 48.3 & 47.3 & 82.2 & 52.7 & 53.3 & 51.7 & 66.3\\
        \midrule
        Qwen3-VL-8B & 56.8 & 67.3 & 69.9 & 84.2 & 71.6 & 87.7 & 71.0 & \underline{77.9} & 86.7 & 84.2 & 59.1 & 51.9 & 98.5 & 71.7 & 79.7 & 66.7 & \underline{86.9}\\
        Qwen3-VL-8B + TrustJudge~\cite{wang2025trustjudge} & 56.8 & 63.4 & 62.7 & 80.1 & 71.2 & 83.9 & 58.6 & 76.0 & 84.4 & 73.6 & 55.9 & 53.2 & 97.9 & 70.8 & 73.6 & 59.0 & 84.7\\
        Qwen3-VL-8B + self-consistency~\cite{wang2023selfconsistency} & 55.5 & 66.7 & \underline{77.8} & 72.3 & 55.3 & 82.2 & \underline{78.5} & 69.7 & 60.0 & 73.1 & 65.0 & \underline{58.9} & 78.2 & 57.0 & 74.0 & \underline{73.8} & 73.4\\
        Qwen3-VL-8B-Thinking~\cite{bai2025qwen3} & 55.5 & \underline{74.1} & \textbf{83.8} & 73.3 & 64.6 & \underline{90.6} & \textbf{88.5} & 72.0 & \textbf{95.1} & \textbf{98.1} & \textbf{79.6} & \textbf{78.5} & \underline{99.6} & 71.7 & \underline{87.6} & \textbf{84.0} & 81.6\\
        UnifiedReward-Think-8B (pair)$^{\dagger}$~\cite{wang2025unifiedrewardthink} & 57.5 & 64.6 & 82.0 & 91.8 & 69.6 & 78.7 & 78.7 & 86.4 & 70.7 & 64.1 & 58.9 & 50.4 & 98.5 & 65.9 & 69.1 & 73.2 & 92.2\\
        VideoSSR-8B~\cite{he2026videossr} & \underline{59.0} & 69.8 & 71.0 & \underline{84.5} & \underline{73.1} & 81.9 & 74.7 & 77.7 & 89.8 & 81.5 & \underline{69.3} & 55.2 & 98.5 & \underline{74.0} & 77.7 & 71.7 & \underline{86.9}\\
        \rowcolor{oursrow}Qwen3-VL-8B + \textbf{Ours} & \textbf{62.9} & \textbf{83.5} & 66.9 & \textbf{91.7} & \textbf{81.8} & \textbf{95.6} & 69.3 & \textbf{85.7} & \underline{92.4} & \underline{96.0} & 68.1 & 52.1 & \textbf{99.7} & \textbf{79.0} & \textbf{91.7} & 68.1 & \textbf{92.4}\\
        \rowcolor{oursrow} & \dcolp{6.1} & \dcolp{16.2} & \dcolm{3.0} & \dcolp{7.5} & \dcolp{10.2} & \dcolp{7.9} & \dcolm{1.7} & \dcolp{7.8} & \dcolp{5.7} & \dcolp{11.8} & \dcolp{9.0} & \dcolp{0.2} & \dcolp{1.2} & \dcolp{7.3} & \dcolp{12.0} & \dcolp{1.4} & \dcolp{5.5}\\[1pt]
        \midrule
        LLaVA-NeXT-Video-7B & \underline{52.4} & 51.9 & \textbf{53.8} & \underline{70.4} & \underline{52.0} & \underline{52.0} & \textbf{58.1} & \underline{62.8} & 48.9 & \underline{51.3} & 49.1 & 48.9 & \textbf{55.7} & \underline{51.1} & \underline{51.8} & \textbf{53.6} & \underline{63.0}\\
        LLaVA-NeXT-Video-7B + self-consistency~\cite{wang2023selfconsistency} & 48.0 & \underline{53.6} & \textbf{53.8} & 62.8 & 49.9 & 51.4 & \underline{55.1} & 56.4 & \underline{49.0} & 48.2 & \textbf{50.5} & \textbf{51.3} & 53.4 & 49.0 & 51.1 & \underline{53.1} & 57.5\\
        \rowcolor{oursrow}LLaVA-NeXT-Video-7B + \textbf{Ours} & \textbf{67.4} & \textbf{64.3} & \underline{50.9} & \textbf{72.5} & \textbf{78.6} & \textbf{67.9} & 52.5 & \textbf{64.4} & \textbf{53.9} & \textbf{51.9} & \underline{50.1} & \underline{50.1} & \underline{53.7} & \textbf{66.6} & \textbf{61.4} & 51.2 & \textbf{63.6}\\
        \rowcolor{oursrow} & \dcolp{15.0} & \dcolp{12.4} & \dcolm{2.9} & \dcolp{2.1} & \dcolp{26.6} & \dcolp{15.9} & \dcolm{5.6} & \dcolp{1.6} & \dcolp{5.0} & \dcolp{0.6} & \dcolp{1.0} & \dcolp{1.2} & \dcolm{2.0} & \dcolp{15.5} & \dcolp{9.6} & \dcolm{2.4} & \dcolp{0.6}\\[1pt]
        \midrule
        InternVL3.5-8B & \underline{58.3} & \underline{71.4} & 63.0 & \underline{83.1} & \underline{74.2} & \underline{92.2} & 62.7 & 78.5 & 92.5 & 92.3 & \underline{70.6} & 68.2 & \textbf{100.0} & \underline{75.0} & \underline{85.3} & 65.4 & \underline{87.2}\\
        InternVL3.5-8B + self-consistency~\cite{wang2023selfconsistency} & 57.4 & 71.2 & \textbf{74.1} & 78.9 & 70.6 & 90.5 & \textbf{78.0} & \underline{81.3} & \underline{93.4} & \underline{92.4} & \textbf{81.5} & \textbf{76.5} & 99.6 & 73.8 & 84.7 & \textbf{77.9} & 86.6\\
        \rowcolor{oursrow}InternVL3.5-8B + \textbf{Ours} & \textbf{66.4} & \textbf{90.0} & \underline{68.7} & \textbf{89.0} & \textbf{86.4} & \textbf{98.4} & \underline{66.3} & \textbf{83.0} & \textbf{99.0} & \textbf{99.8} & 69.4 & \underline{68.9} & \underline{99.9} & \textbf{84.0} & \textbf{96.1} & \underline{68.1} & \textbf{90.6}\\
        \rowcolor{oursrow} & \dcolp{8.1} & \dcolp{18.6} & \dcolp{5.7} & \dcolp{5.9} & \dcolp{12.2} & \dcolp{6.2} & \dcolp{3.6} & \dcolp{4.5} & \dcolp{6.5} & \dcolp{7.5} & \dcolm{1.2} & \dcolp{0.7} & \dcolm{0.1} & \dcolp{9.0} & \dcolp{10.8} & \dcolp{2.7} & \dcolp{3.4}\\[1pt]
        \bottomrule
    \end{tabular}}
\end{table*}

%% file: sections/5_experiments.tex
\section{Experiments}
\label{sec:Experiments}

To address the weak order preferences and shared offsets (\Cref{sec:Diagnosing_Temporal_Judgment}), \method learns from within-video comparisons. It consistently boosts order discrimination across settings while the content anchor preserves content.

% ----------------------------------------------------------------
% 5.1 Experimental Setup
% ----------------------------------------------------------------
\subsection{Experimental Setup}
\label{sec:Experimental_Setup}

\paragraph{Data and metrics.}
We construct training data from 1,493 source videos in Charades~\cite{sigurdsson2016charades} and CLEVRER~\cite{yi2020clevrer} without human-written claims or manual order labels (\Cref{tab:Judge_sets}).
Temporal order labels follow the chronological positions of sampled segments.
The frozen base model describes each segment and selects records by segment-description correspondence.
After description checks and selection, 1,933 training records remain for Qwen3-VL-8B.
We evaluate on three judge sets with video-disjoint splits: NExT-GQA-Judge (NG-J)~\cite{xiao2024nextgqa}, Charades-Judge (Ch-J), and CLEVRER-Judge (Cl-J).
Each test item contains one chronological claim, five order negatives, and one content negative where valid.
We report AUC $\times 100$.
Pooled AUC measures discrimination across the full test set, whereas within-video AUC restricts comparisons to claims from the same video.
We use pooled AUC to distinguish improvements in score discrimination from mere shifts in the \tokYes{}/\tokNo{} decision threshold.
Complementary within-video ranking metrics and accuracy
results are provided in~\Cref{app:Analysis_of_Judgment_Improvements}.

\vspace{-0.5em}
\paragraph{Models and baselines.}
We apply \method to Qwen3-VL-8B-Instruct~\cite{bai2025qwen3}, LLaVA-NeXT-Video-7B~\cite{zhang2024llavanextvideo}, and InternVL3.5-8B~\cite{wang2025internvl35}.
We train a LoRA adapter~\cite{hu2021lora} on first-round records generated and selected by each backbone.
Training compares claims within each video, whereas single-claim inference uses one pass to obtain the score $z_{\mathrm{Yes}}-z_{\mathrm{No}}$.
Baselines include released multimodal judges, two inference-time readout adaptations, and Qwen3-VL-8B-Thinking.
We also compare with VideoSSR~\cite{he2026videossr} on the same Qwen backbone.
Details are in \Cref{app:Implementation_Details,app:Baseline_Adaptations_and_Inference_Cost}.

% ----------------------------------------------------------------
% 5.2 Main Results
% ----------------------------------------------------------------
\subsection{Main Results}
\label{sec:Main_Results}

\method improves pooled order AUC on every judge set across all three backbones (\Cref{tab:Temporal_judgment_and_content_discrimination}).
The average gains are 7.3 points on Qwen3-VL-8B, 15.5 on LLaVA-NeXT-Video-7B, and 9.0 on InternVL3.5-8B.
These gains also extend to NG-J, which provides no training videos.

On Qwen3-VL-8B, content AUC also improves on all three sets by 5.5 points on average.
Content AUC improves or remains near its ceiling on InternVL3.5-8B.
On Qwen3-VL-8B, \method exceeds VideoSSR in both order and content AUC on all three sets.
It also exceeds both inference-time readout adaptations in average order AUC.
These gains show that within-video training improves order discrimination
under single-claim inference.

% ----------------------------------------------------------------
% 5.3 Ablation Study
% ----------------------------------------------------------------
\subsection{Ablation Study}
\label{sec:Ablation_Study}

\input{tables/tab_3_ablations}

\Cref{tab:Ablations} evaluates the contributions of chronological supervision and loss components on Qwen3-VL-8B.
Correct chronological labels yield an average order AUC of 79.0, compared with 49.7 under the random-label control.
The control still achieves 91.6 content AUC.
This contrast shows that high content discrimination alone does not ensure reliable order judgment without valid chronological supervision.
The no-video regularizer in $\mathcal L_{\mathrm{aux},i}$ discourages text-only order preferences and improves order AUC by 1.7.

The content anchor raises average content AUC from 87.9 to 92.4.
It penalizes increases in content-negative scores relative to the frozen base judge.
This constraint trades some order discrimination for stronger content discrimination.
Average order AUC is 79.0 with the anchor and 83.9 without it.
We retain the anchor for its joint gains of 7.3 order points and 5.5 content points over the baseline.
Additional ablation studies are in \Cref{app:Additional_Ablations}.

\input{figures/fig_5_offset_after}

% ----------------------------------------------------------------
% 5.4 Analysis of Judgment Improvements
% ----------------------------------------------------------------
\input{figures/fig_4_qualitative}

\subsection{Analysis of Judgment Improvements}
\label{sec:Analysis_of_Judgment_Improvements}

On Qwen3-VL-8B, training strengthens within-video order preferences
relative to cross-video score variation.
We revisit the signed order gap $\delta_i$ and shared score offset $b_i$
from \Cref{sec:Order_Sensitivity_and_Score_Offsets}.
\Cref{fig:Score_decomposition_before_and_after_training} compares the
model before and after training under the same judgment prompt.
The paired scores in
\Cref{fig:Score_decomposition_before_and_after_training}\textcolor{cvprblue}{(a)} show larger
gaps between chronological and reordered claims.

The ranking loss cancels $b_i$ within each comparison but does not
directly minimize cross-video offset variation.
Offsets persist at inference, and their spread increases after training.
The mean order gap nevertheless grows relative to the offset standard
deviation on all three validation sets
as in \Cref{fig:Score_decomposition_before_and_after_training}\textcolor{cvprblue}{(b)}.
On the test sets, pooled order AUC improves by 7.3 points on average,
compared with 1.9 points for within-video order AUC
(\Cref{app:Analysis_of_Judgment_Improvements}).
Together, the stronger order preferences relative to offset variation
and the larger pooled gains support improved cross-video order
discrimination under single-claim inference.

% ----------------------------------------------------------------
% 5.5 Scope of Generalization
% ----------------------------------------------------------------

\subsection{Scope of Generalization}
\label{sec:Scope_of_Generalization}

\input{tables/tab_4_external_judge}

\paragraph{Temporal localization.}
The gains extend from event order to coarse temporal localization
(\Cref{tab:Temporal_judgment_and_content_discrimination}).
This task evaluates claims that place an event in the first or second half
of a video.
Localization AUC improves on all three judge sets for every backbone.
The average gains are 12.0 points on Qwen3-VL-8B, 9.6 on
LLaVA-NeXT-Video-7B, and 10.8 on InternVL3.5-8B.
These gains show that order supervision transfers to judgments of temporal
position without localization claims in training.
Our method also improves duration judgment in some settings, while gains
in event counting remain limited.
Unlike event order, these tasks require estimating temporal extent or
event frequency, which our order contrasts do not directly supervise.
Additional claim formats and temporal attributes are in
\Cref{app:Transfer_to_Other_Temporal_Claims}.
\vspace{-1.2em}
\paragraph{External benchmarks.}
\method improves pooled order AUC across all three backbones on the external
human-verified benchmarks TempCompass~\cite{liu2024tempcompass},
VELOCITI~\cite{saravanan2025velociti}, and
Vinoground~\cite{zhang2024vinoground}
(\Cref{tab:External_judgment_benchmarks}).
For Qwen3-VL-8B, the gains range from 4.7 to 10.8 points.
These results show that learning from within-video comparisons improves
order discrimination.

% ----------------------------------------------------------------
% 5.6 Qualitative Results
% ----------------------------------------------------------------
\subsection{Qualitative Results}
\label{sec:Qualitative_Results}

\Cref{fig:Selected_corrected_judgments} illustrates examples of standalone
temporal judgments before and after training.
In (a), both claims describe the same events in different orders.
The base model ranks the chronological claim higher but accepts both claims.
\method accepts the chronological claim and rejects its reordered
counterpart.
This illustrates that correct relative ranking does not necessarily imply
correct standalone verdicts.
Each claim is scored independently, so this correction is achieved without
alternative candidates at inference.
Panel (b) shows corrected temporal localization judgments.
Panels (c) and (d) illustrate selected corrections of duration and frequency
claims.

%% file: tables/tab_3_ablations.tex
\begin{table}[t]
\centering
\footnotesize
\setlength{\tabcolsep}{2pt}
\caption{
    \textbf{Ablations.} Pooled test AUC $\times100$ on Qwen3-VL-8B.
    Round 2 re-selects the records with the round-1 judge and the base generator (\Cref{sec:Iterating_the_Self_Improving_Loop}).
    rand.: random order labels.
}
\label{tab:Ablations}
\resizebox{\columnwidth}{!}{%
\begin{tabular}{
    @{}>{\hspace{1.2pt}}c<{\hspace{1.2pt}}>{\hspace{1.2pt}}c<{\hspace{1.2pt}}>{\hspace{1.2pt}}c<{\hspace{1.2pt}}>{\hspace{1.2pt}}c<{\hspace{1.2pt}}cccccccc@{}
}
    \toprule
    \raisebox{-13pt}[\dimexpr\height-13pt\relax][0pt]{\rotatebox[origin=l]{90}{Order}} & \raisebox{-13pt}[\dimexpr\height-13pt\relax][0pt]{\rotatebox[origin=l]{90}{Anchor}} & \raisebox{-13pt}[\dimexpr\height-13pt\relax][0pt]{\rotatebox[origin=l]{90}{No-video}} & \raisebox{-13pt}[\dimexpr\height-13pt\relax][0pt]{\rotatebox[origin=l]{90}{Rounds}} & \multicolumn{2}{c}{NG-J} & \multicolumn{2}{c}{Ch-J} & \multicolumn{2}{c}{Cl-J} & \multicolumn{2}{c}{\textit{Average}}\\
    \cmidrule(lr){5-6}\cmidrule(lr){7-8}\cmidrule(lr){9-10}\cmidrule(lr){11-12}
     &  &  &  & Ord. & Content & Ord. & Content & Ord. & Content & Ord. & Content\\
    \midrule
    \multicolumn{4}{l}{Base} & 56.8 & 84.2 & 71.6 & 77.9 & 86.7 & 98.5 & 71.7 & 86.9\\
    \midrule
    rand. & $\checkmark$ & $\checkmark$ & 1 & 50.3 & 91.0 & 51.2 & 83.8 & 47.6 & \textbf{100.0} & 49.7 & 91.6\\
     &  &  &  & \dcolm{6.5} & \dcolp{6.8} & \dcolm{20.4} & \dcolp{5.9} & \dcolm{39.1} & \dcolp{1.5} & \dcolm{22.0} & \dcolp{4.7}\\[1pt]
    $\checkmark$ & $\times$ & $\checkmark$ & 1 & \textbf{70.7} & 87.5 & \textbf{85.4} & 77.6 & \textbf{95.6} & 98.6 & \textbf{83.9} & 87.9\\
     &  &  &  & \dcolp{13.9} & \dcolp{3.3} & \dcolp{13.8} & \dcolm{0.3} & \dcolp{8.9} & \dcolp{0.1} & \dcolp{12.2} & \dcolp{1.0}\\[1pt]
    $\checkmark$ & $\checkmark$ & $\times$ & 1 & 60.9 & 91.1 & 79.8 & 84.1 & 91.2 & 99.8 & 77.3 & 91.7\\
     &  &  &  & \dcolp{4.1} & \dcolp{6.9} & \dcolp{8.2} & \dcolp{6.2} & \dcolp{4.5} & \dcolp{1.3} & \dcolp{5.6} & \dcolp{4.8}\\[1pt]
    \midrule
    $\checkmark$ & $\checkmark$ & $\checkmark$ & 2 & 63.1 & 91.6 & 81.7 & 83.7 & 95.1 & 99.9 & 79.9 & 91.7\\
     &  &  &  & \dcolp{6.3} & \dcolp{7.4} & \dcolp{10.1} & \dcolp{5.8} & \dcolp{8.4} & \dcolp{1.4} & \dcolp{8.2} & \dcolp{4.8}\\[1pt]
    \midrule
    \rowcolor{oursrow}$\checkmark$ & $\checkmark$ & $\checkmark$ & 1 & 62.9 & \textbf{91.7} & 81.8 & \textbf{85.7} & 92.4 & 99.7 & 79.0 & \textbf{92.4}\\
    \rowcolor{oursrow} &  &  &  & \dcolp{6.1} & \dcolp{7.5} & \dcolp{10.2} & \dcolp{7.8} & \dcolp{5.7} & \dcolp{1.2} & \dcolp{7.3} & \dcolp{5.5}\\[1pt]
    \bottomrule
\end{tabular}}
\end{table}

%% file: figures/fig_5_offset_after.tex
\begin{figure}[t]
\vspace{-1em}
\centering
\includegraphics[width=\linewidth]{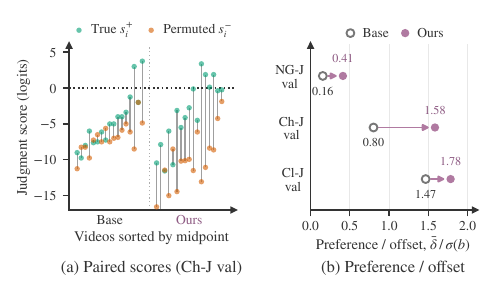}
\vspace{-2em}
\caption{
    \textbf{Score decomposition before and after training.}
    (a) Paired scores on Ch-J validation videos for Qwen3-VL-8B show that training widens within-video order gaps while cross-video offsets $b_i$ persist.
    (b) The ratio of mean order preference to offset spread ($\bar{\delta}/\sigma(b)$, with $b$ and $\delta$ from \Cref{eq:decomposition}) rises across all three validation sets.
}
\label{fig:Score_decomposition_before_and_after_training}
\vspace{-0.5em}
\end{figure}

%% file: figures/fig_4_qualitative.tex
\begin{figure*}[bt]
    \centering
    \includegraphics[width=\textwidth]{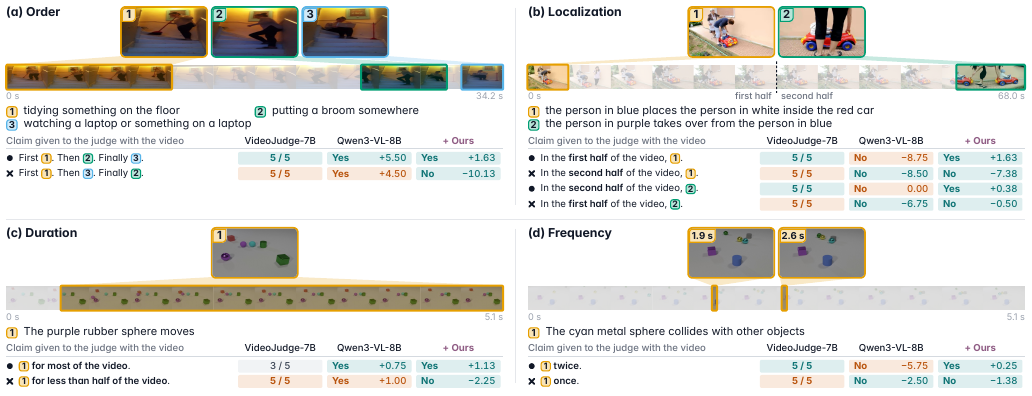}
    \caption{
        \textbf{Examples of temporal judgments.}
        (a) Event order (Ch-J).
        (b) Coarse temporal localization (NG-J).
        (c) Duration and (d) frequency (Cl-J), where aggregate improvements are limited or inconsistent.
        $\bullet$ and $\times$ indicate true and false claims.
        Cells show verdicts and \tokYes{} minus \tokNo{} logits for Qwen3-VL-8B before and after training.
        VideoJudge-7B uses its official 1--5 ratings, where 4--5 indicate acceptance, 1--2 rejection, and 3 is intermediate.
        Additional details are provided in \Cref{app:Qualitative_Examples}.
    }
    \label{fig:Selected_corrected_judgments}
\end{figure*}

%% file: tables/tab_4_external_judge.tex
\begin{table}[t]
    \centering
    \footnotesize
    \setlength{\tabcolsep}{2.6pt}
    \caption{
        \textbf{External judgment benchmarks.} Pooled AUC ($\times$100) on TempCompass~\cite{liu2024tempcompass}, VELOCITI~\cite{saravanan2025velociti}, and Vinoground~\cite{zhang2024vinoground} (details in \Cref{app:External_Benchmarks_Video_QA_and_Structural_Transfer}). Parentheses: change from the base model.
    }
    \vspace{-0.5em}
    \label{tab:External_judgment_benchmarks}
    \begin{tabular}{lccccc}
        \toprule
         & \multicolumn{2}{c}{TempCompass} & \multicolumn{2}{c}{VELOCITI} & Vinoground\\
        \cmidrule(lr){2-3}\cmidrule(lr){4-5}\cmidrule(lr){6-6}
        Model & Order & Others & Order & Others & Order\\
        \midrule
        Qwen3-VL-8B & 88.9 & 84.9 & 68.0 & 81.6 & 65.3\\
        \rowcolor{oursrow}\quad + \textbf{Ours} & \vdc{93.6}{\dplus{4.7}} & \vdc{85.1}{\dplus{0.2}} & \vdc{76.5}{\dplus{8.5}} & \vdc{79.9}{\dminus{1.6}} & \vdc{76.2}{\dplus{10.8}}\\
        \midrule
        LLaVA-NeXT-Video-7B & 55.1 & 63.6 & 51.6 & 60.0 & 50.8\\
        \rowcolor{oursrow}\quad + \textbf{Ours} & \vdc{76.8}{\dplus{21.7}} & \vdc{65.6}{\dplus{2.0}} & \vdc{83.8}{\dplus{32.2}} & \vdc{56.3}{\dminus{3.7}} & \vdc{67.1}{\dplus{16.3}}\\
        \midrule
        InternVL3.5-8B & 85.5 & 80.9 & 71.2 & 79.0 & 68.1\\
        \rowcolor{oursrow}\quad + \textbf{Ours} & \vdc{92.5}{\dplus{6.9}} & \vdc{83.5}{\dplus{2.5}} & \vdc{90.5}{\dplus{19.3}} & \vdc{81.5}{\dplus{2.6}} & \vdc{80.8}{\dplus{12.7}}\\
        \bottomrule
    \end{tabular}
    \vspace{-1em}
\end{table}

%% file: sections/7_conclusion.tex
\Needspace{6\baselineskip}
\section{Conclusion}
\label{sec:Conclusion}

We identify a discrepancy between temporal understanding and standalone judgment in video LLMs, where temporal information is not reliably reflected in judgment scores.
Our analysis reveals weak temporal order discrimination and shared score variation across videos.
To address this discrepancy, we propose \method, a framework that learns from self-generated, content-preserving comparisons while retaining single-claim inference.
Experiments demonstrate consistent improvements in temporal order judgment across video LLMs and benchmarks, with generalization to coarse temporal localization.
These findings highlight the importance of bridging the gap between temporal understanding and reliable judgment in video LLMs.
We discuss limitations and directions for future work in \Cref{app:Limitations_and_Future_Work}.

%% file: sections/8_llm_usage.tex
\section*{LLM Usage Statement}
GPT~\cite{openai2025gpt41} and Claude~\cite{anthropic2025claude} were used to assist with the validation of evaluation items, as detailed in the supplementary material.
Claude Code and OpenAI Codex were also used to assist with code implementation and manuscript polishing.
The authors designed the study, verified the implementation and results, and take full responsibility for the content of this paper.

%% file: sections/app_0_contents.tex
\begingroup
\phantomsection
\pdfbookmark[0]{Appendix}{appendix-overview}
\small
\hypersetup{linktoc=all}
\etocsettagdepth{main}{none}
\etocsettagdepth{appendix}{subsection}
\etocsettocstyle{\section*{Contents}}{}
\makeatletter
\newcommand{\appendixtocname}[1]{%
    \def\appendixtoctitle{#1}%
    \patchcmd{\appendixtoctitle}{\hskip -1em.\nobreakspace {}}{}{}{}%
    \appendixtoctitle}
\renewcommand{\l@section}[2]{%  
    \addvspace{5pt}%
    \@dottedtocline{1}{0em}{1.8em}{\bfseries\appendixtocname{#1}}{\bfseries #2}}
\renewcommand{\l@subsection}[2]{%
    \@dottedtocline{2}{1.8em}{2.8em}{\appendixtocname{#1}}{#2}}
\makeatother
\tableofcontents

\endgroup

\newpage

\section*{Overview of the Supplementary Material}
The supplementary material begins with limitations, future directions, and extended related work (\Cref{app:Limitations_and_Future_Work,app:Extended_Related_Work}).
It then provides detailed evaluation protocols, data construction, implementation details, and baseline settings (\Cref{app:Judge_Evaluation_Sets,app:Metrics_and_Evaluation_Protocol,app:Notation,app:Data_Construction_and_Semantic_Validity,app:Implementation_Details,app:Prompts,app:Baseline_Adaptations_and_Inference_Cost}).
The remaining sections present extended diagnostics, additional experiments and ablations, analyses of judgment improvements and generalization, and qualitative examples (\Cref{app:Extended_Diagnostics,app:Additional_Backbones,app:Additional_Ablations,app:Analysis_of_Judgment_Improvements,app:Transfer_to_Other_Temporal_Claims,app:External_Benchmarks_Video_QA_and_Structural_Transfer,app:Qualitative_Examples}).
Throughout the supplementary material, Ours denotes \method, the framework proposed in the main paper.
Code is available at {\urlstyle{same}\url{https://github.com/kaist-cvml/temporal-judge}}.

%% file: sections/app_0_main.tex
\nocite{yang2025qwen3}

% Discussion

\input{sections/app_A_limitations}

\input{sections/app_B_related_work}

% Setup: evaluation (C, D), method (E-H), baselines (I)

\input{sections/app_C_judge_sets}

\input{sections/app_D_metrics}

\input{sections/app_E_notation}

\input{sections/app_F_data_construction}

\input{sections/app_G_implementation}

\input{sections/app_H_prompts}

\input{sections/app_I_baselines}

% Results, in the order of the main paper: Section 3 (J), 5.2 (K), 5.3 (L), 5.4 (M), 5.5 (N, O)
\input{sections/app_J_diagnostics}

\input{sections/app_K_backbones}

\input{sections/app_L_ablations}

\input{sections/app_M_analysis}

\input{sections/app_N_transfer}

\input{sections/app_O_external}

% Qualitative examples

\input{sections/app_P_qualitative}

%% file: sections/app_A_limitations.tex
\section{Limitations and Future Work}
\label{app:Limitations_and_Future_Work}

The training claims and their order labels are generated without manual annotation.
The order labels follow the chronological positions of the sampled segments, and the base model generates the event descriptions.
A description can therefore be inaccurate or can also match another segment, so some training contrasts are noisy.
Confidence-based selection filters records by segment-description correspondence but does not guarantee semantic correctness (\Cref{app:Data_Construction_and_Semantic_Validity}).
Evaluation labels primarily derive from source annotations, with NG-J duration as a model-assisted exception; construction filters reduce ambiguity from repeated or overlapping events (\Cref{app:Judge_Evaluation_Sets}).
Independent human validation could further strengthen the reliability of these labels.

\paragraph{What the analysis identifies.}
Our probing and score analyses characterize the discrepancy between temporal understanding and judgment without establishing its causal mechanism (\Cref{sec:Order_Sensitivity_and_Score_Offsets}).
Training strengthens temporal order preferences relative to cross-video score variation (\Cref{sec:Analysis_of_Judgment_Improvements}).
As our primary metrics measure score discrimination, further improving the calibration of individual judgments remains a direction for future work.

\paragraph{Future work.}
The limited transfer to duration, frequency, and event overlap motivates comparisons that directly supervise these attributes, potentially combined with reasoning-based video LLMs (\Cref{tab:Temporal_judgment_and_content_discrimination}).
Another direction is to improve the quality of generated comparisons and their selection across rounds.
These extensions could broaden the use of temporal judges in evaluating video generation and world models~\cite{nvidia2025cosmos,zheng2025vbench2}.

%% file: sections/app_B_related_work.tex
\section{Extended Related Work}
\label{app:Extended_Related_Work}

\paragraph{Multimodal LLM-as-a-Judge.}
MLLM-as-a-Judge~\cite{chen2024mllmjudge} investigates multimodal judging capabilities and reliability across different evaluation tasks.
Prior work has developed learned evaluators, including Prometheus-Vision~\cite{lee2024prometheusvision}, Flex-Judge~\cite{ko2025flexjudge}, Prometheus~2~\cite{kim2024prometheus2}, and JudgeLM~\cite{zhu2025judgelm}.
Recent studies further address perceptual judgment biases~\cite{park2026perceptionjudge} and self-improving VLM judges trained with synthetic supervision~\cite{lin2025selfjudges}.
For video evaluation, VideoJudge~\cite{waheed2025videojudge} trains video-conditioned evaluators through bootstrapping, while TimeJudge~\cite{hu2025timejudge} assesses temporal consistency through auxiliary verification questions and answer aggregation.
These approaches advance multimodal evaluation through general-purpose judgment, bias mitigation, or specialized verification.
Building on these advances, our framework focuses on improving temporal order discrimination through self-generated, content-preserving comparisons while retaining independent single-claim inference.

\paragraph{Judgment Reliability and Evaluation Protocols.}
Prior work has shown that LLM-as-a-Judge reliability depends on scoring methods and evaluation protocols~\cite{li2025generationjudgment}.
G-Eval employs probability-weighted ratings~\cite{liu2023g}, while TrustJudge combines distribution-sensitive scoring with pairwise aggregation~\cite{wang2025trustjudge}.
Beyond score computation, the evaluation protocol determines whether claims are assessed independently or through direct comparisons.
Pairwise and pointwise evaluation can exhibit different biases~\cite{tripathi2025pairwise}.
Relatedly, our controlled analysis examines how different task formats affect temporal judgment when evaluating the same underlying temporal fact (\Cref{sec:Diagnosing_Temporal_Judgment}).

\paragraph{Probing what models encode.}
Prior studies show that model representations can encode information not reliably expressed in their outputs~\cite{burns2023latent}, and that access to relevant visual evidence does not always ensure correct predictions~\cite{liu2026seeing}.
We investigate this discrepancy in temporal judgment by probing frozen representations with claims that share the same event descriptions and template but differ in temporal order.
This controlled construction reduces lexical and structural differences between positive and negative claims, helping isolate temporal information in the representations.
Additional probing results, including a label-shuffled control, are reported in \Cref{app:Extended_Diagnostics}.

\paragraph{Video benchmarks.}
Video-MME evaluates general video understanding capabilities~\cite{fu2025video}, while prior studies highlight challenges in temporal ordering and moment grounding~\cite{xiao2025videoqa}.
MomentMix investigates temporal robustness in moment retrieval through video augmentation~\cite{park2026momentmix}.
These studies provide complementary perspectives on temporal video understanding.
Our evaluation instead focuses on standalone judgment of temporal relations using controlled claims constructed from annotated events (\Cref{app:Judge_Evaluation_Sets}).
By preserving event descriptions across chronological and reordered claims, we assess temporal discrimination separately from event-content recognition.

\input{tables/tab_C_judge_stats}

\paragraph{Contrast and self-generated negatives.}
Prior contrastive approaches include inference-time visual contrastive decoding~\cite{leng2024mitigating}, training with hallucinated captions~\cite{jiang2024hacl}, and confidence calibration using self-generated distractors~\cite{wang2026calibrating}.
Our order negatives instead permute model-generated event descriptions while preserving their content, isolating temporal order from event content.
The resulting within-video comparisons directly supervise temporal order preferences, while a content anchor helps preserve event-content discrimination.
Unlike inference-time contrastive methods, our framework learns these preferences during training and independently evaluates each video-claim pair without requiring alternative claims.

%% file: tables/tab_C_judge_stats.tex
\begin{table*}[t]
\centering
\small
\setlength{\tabcolsep}{5pt}
\caption{
    \textbf{Statistics of the test sets.}
    Order evaluation uses the chronological claim and its five alternative orderings.
    Content evaluation uses videos with a retained content negative.
    Each localization, duration, or frequency instance yields one true and one false claim.
    The Cl-J duration set also includes videos from the CLEVRER validation split that are outside the Cl-J pool and were excluded from training.
}
\label{tab:Judge_set_statistics}
\begin{tabular}{lcccccccccc}
    \toprule
    & \multicolumn{3}{c}{Order} & \multicolumn{3}{c}{Localization} & \multicolumn{3}{c}{Duration} & Frequency\\
    \cmidrule(lr){2-4}\cmidrule(lr){5-7}\cmidrule(lr){8-10}\cmidrule(lr){11-11}
     & NG-J & Ch-J & Cl-J & NG-J & Ch-J & Cl-J & NG-J & Ch-J & Cl-J & Cl-J\\
    \midrule
    Videos & 305 & 330 & 331 & 86 & 265 & 324 & 60 & 157 & 455 & 330\\
    Claims & 1,830 & 1,980 & 1,986 & 192 & 592 & 956 & 128 & 356 & 1,316 & 1,980\\
    Video length, median (s) & 41.1 & 31.2 & 5.12 & 34.5 & 31.5 & 5.12 & 45.5 & 31.7 & 5.12 & 5.12\\
    Words per claim, median & 29 & 15 & 33 & 15 & 14 & 17 & 14 & 13 & 11 & 9\\
    \bottomrule
\end{tabular}%
\end{table*}

%% file: sections/app_C_judge_sets.tex
\input{figures/fig_C_order_examples}

\input{figures/fig_C_localization_claims}

\input{figures/fig_C_duration_claims}

\input{figures/fig_C_frequency_claims}

\section{Judge Evaluation Sets}
\label{app:Judge_Evaluation_Sets}

We construct three judge evaluation sets from existing temporally annotated datasets: NExT-GQA-Judge (NG-J), Charades-Judge (Ch-J), and CLEVRER-Judge (Cl-J).
Each evaluation item contains a chronological claim describing three events (``First $e_1$. Then $e_2$. Finally $e_3$.''), five alternative orderings, and a candidate content negative formed by replacing one event with an event from another video.
Order negatives preserve event content while varying temporal order, whereas content negatives assess event-content discrimination.
\Cref{tab:Judge_set_statistics} summarizes the evaluation sets, \Cref{fig:Order_and_content_claims} provides examples, and \Cref{app:Metrics_and_Evaluation_Protocol} defines the metrics.

Our main evaluation uses three-event claims, with additional experiments covering different event counts and templates (\Cref{app:Event_count_and_template}) and other temporal attributes (\Cref{app:Temporal_attribute_claims}).

\paragraph{Label sources.}
Order labels and most temporal-attribute labels derive from source-dataset annotations.
NG-J duration is an exception: its labels are assigned through agreement between auxiliary visual assessments, because question-evidence spans do not specify event duration.
GPT-4.1~\cite{openai2025gpt41} (GPT) rephrases annotated question-answer pairs into event descriptions (\Cref{app:NExT_GQA_Judge}), and GPT and Claude Opus 5.5~\cite{anthropic2025claude} (Claude) perform the visual checks described in \Cref{app:Temporal_attribute_claims}.
Both auxiliary models are separate from the judges under evaluation.

\subsection{Order and content claims}
\label{app:Order_and_content_claims}

\paragraph{Common construction.}
We select the first three events forming a temporal chain, requiring at least 0.5\,s of separation between consecutive events.
Dataset-specific filtering further reduces ambiguous or highly predictable event sequences and screens content negatives for overlapping event content.
All splits are video-disjoint, with validation and test videos excluded from the training source pool (\Cref{tab:Judge_sets}).
Source annotations guide video selection but are not used as training targets.

\paragraph{NExT-GQA-Judge.}
\label{app:NExT_GQA_Judge}
NG-J is derived from NExT-GQA~\cite{xiao2024nextgqa}, which provides question-answer pairs grounded in temporal evidence spans.
We treat each grounded pair as an event and its evidence span as the event's temporal location.
The test set contains 305 videos, while 211 disjoint validation videos are used for the diagnostics in \Cref{sec:Diagnosing_Temporal_Judgment}.
GPT converts the source question-answer pairs into concise event descriptions without access to the video.
A separate semantic consistency check filters descriptions that do not preserve the original meaning or introduce temporal cues.
This step only paraphrases annotated text, so chronological labels remain determined by the source annotations.

\paragraph{Charades-Judge.}
\label{app:Charades_Judge}
Ch-J uses annotated action instances from Charades~\cite{sigurdsson2016charades}, represented directly by their action-class descriptions without model-based rewriting.
We apply additional filtering to reduce ambiguity from prolonged or repeated actions and to exclude action pairs with strongly predictable chronological order.
The evaluation preserves the official Charades train/test separation, with disjoint source and validation subsets.
Content negatives that also describe annotated events in the target video are excluded from content evaluation.

\paragraph{CLEVRER-Judge.}
\label{app:CLEVRER_Judge}
Cl-J uses collision annotations from CLEVRER~\cite{yi2020clevrer}, where events are described using object attributes such as color, material, and shape.
Collision timestamps provide chronological labels, and videos are selected using the common temporal separation criterion.
We partition the retained videos from the CLEVRER validation split into disjoint source, validation, and test subsets.
The resulting evaluation covers short synthetic videos with precisely annotated collision events.

\subsection{Statistics and text-only control}
\label{app:Statistics_and_text_only_control}

The three sets cover complementary evaluation conditions (\Cref{tab:Judge_set_statistics}).
NG-J contains longer videos with diverse event descriptions, Ch-J uses a structured vocabulary of human actions, and Cl-J consists of short synthetic videos with object-level collision events.

\paragraph{Text-only control.}
Although order negatives preserve event descriptions, language priors may favor particular sequences.
We therefore evaluate the base judge without video frames.
Text-only order AUCs are 52.8, 55.8, and 53.7 on NG-J, Ch-J, and Cl-J, respectively, compared with 56.8, 71.6, and 86.7 with video.
These results indicate limited text-only order discrimination, with larger gains from video conditioning on Ch-J and Cl-J.

\subsection{Temporal attribute claims}
\label{app:Temporal_attribute_claims}

Beyond event order, we construct matched claims targeting temporal localization, duration, and frequency.
Each positive-negative pair preserves event content while changing the temporal attribute under evaluation.
\Cref{fig:Localization_claims,fig:Duration_claims,fig:Frequency_claims} provide examples, and \Cref{tab:Judge_set_statistics} reports the statistics.

\paragraph{Claim construction.}
\emph{Localization} distinguishes whether an event occurs in the first or second half of a video.
\emph{Duration} distinguishes events occurring for most of the video from those lasting less than half of it.
\emph{Frequency} evaluates the number of collisions involving an object and is constructed from CLEVRER annotations.
We retain events with clearly defined temporal attributes and balance positive and negative claims across temporal expressions to reduce textual priors.
For Charades duration claims, we additionally balance duration labels within action classes.
All evaluation videos remain disjoint from training.

\paragraph{Label verification.}
Human annotations can be incomplete, for example when an action continues beyond its annotated span.
We therefore apply visual verification to the localization and duration claims from NG-J and Ch-J.
Claude and GPT independently view the same frames as the judge and answer a coarse question about a single event without access to the reference label, namely in which half of the video the event is visible or whether it covers most or little of the video.
For annotation-derived labels, we retain events when the visual checks support the reference annotation and discard ambiguous cases.
For NG-J duration, the checks instead determine the label, with agreement across all checks required.
CLEVRER labels derive from simulator annotations, and on a sample of CLEVRER events the same label-blind check agrees with the simulator in 182 of 185 decided cases.

\paragraph{Final sets.}
The seven additional temporal-attribute sets contain 5,520 claims across 1,091 videos (\Cref{tab:Judge_set_statistics}), providing targeted evaluations of temporal properties beyond the order-based training objective.

%% file: figures/fig_C_order_examples.tex
\begin{figure*}[t]
\centering
\includegraphics[width=\textwidth]{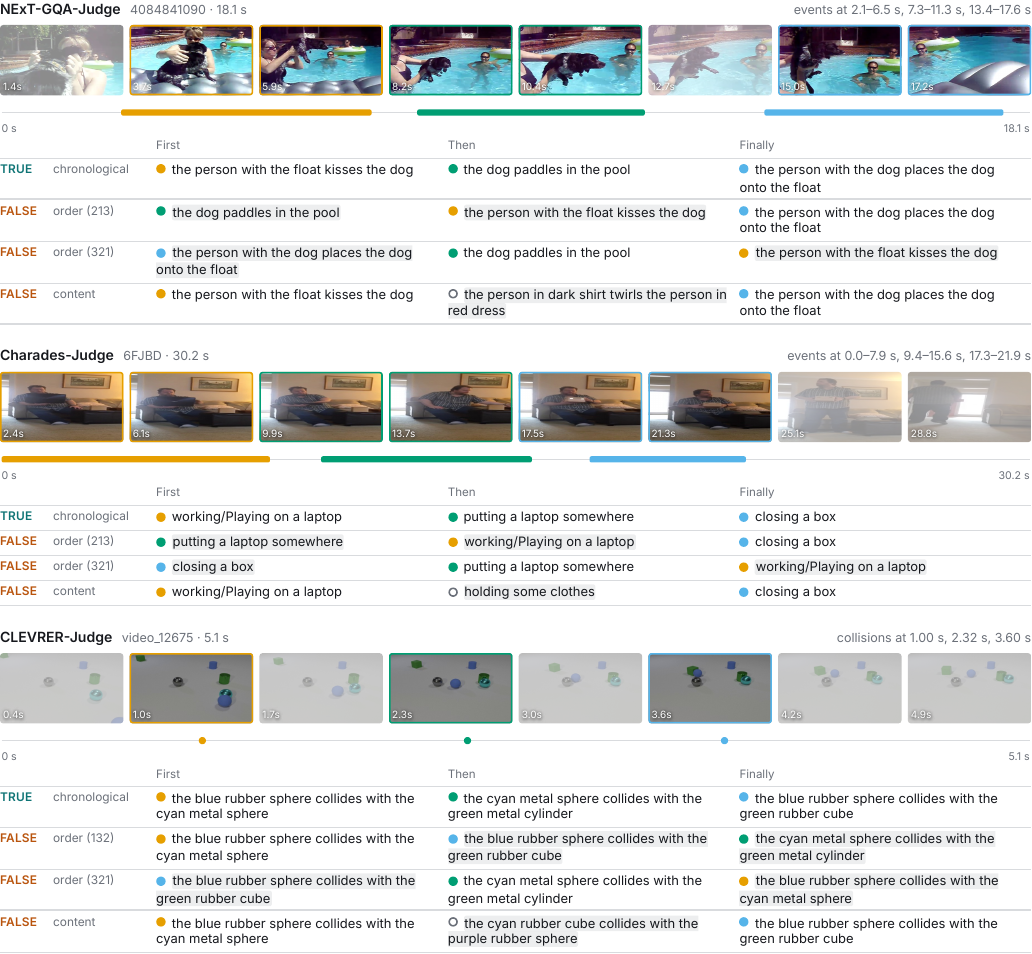}
\caption{
    \textbf{Order and content claims.}
    Examples from the three judge sets.
    Order negatives preserve event content while changing temporal order, whereas content negatives replace an event with one from another video.
    Annotated event intervals are highlighted along the video timelines.
}
\label{fig:Order_and_content_claims}
\end{figure*}

%% file: figures/fig_C_localization_claims.tex
\begin{figure*}[t]
\centering
\includegraphics[width=\textwidth]{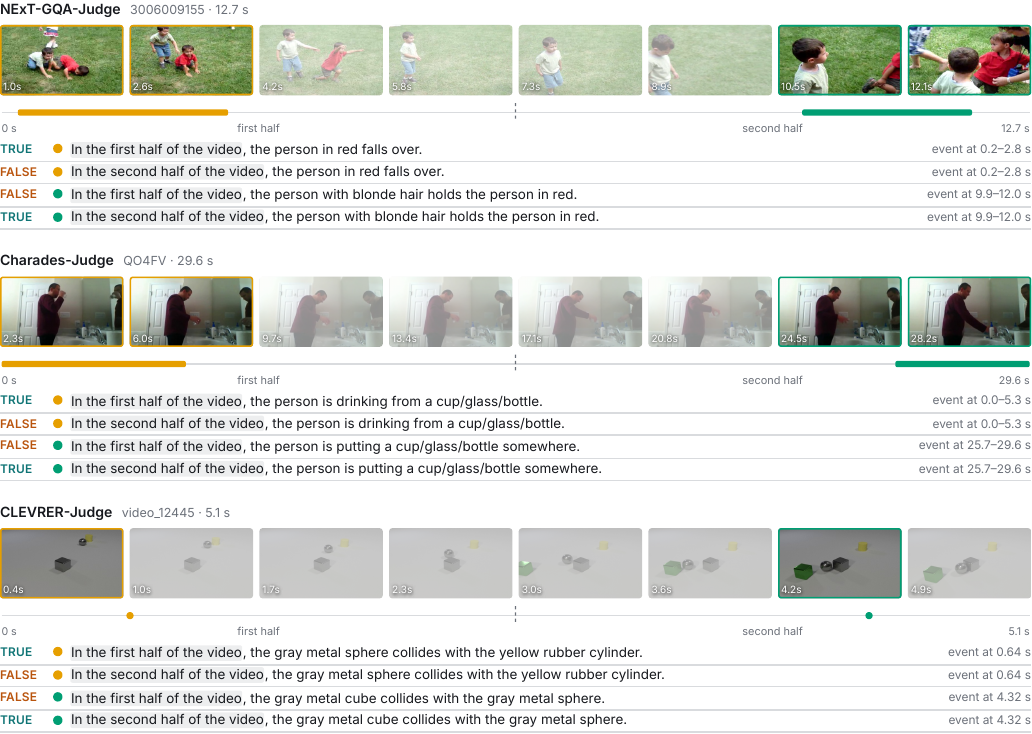}
\caption{
    \textbf{Localization claims.}
    Examples from the three judge sets.
    Positive and negative claims describe the same event but differ in whether it occurs in the first or second half of the video.
    Annotated event intervals are highlighted along the video timelines.
}
\label{fig:Localization_claims}
\end{figure*}

%% file: figures/fig_C_duration_claims.tex
\begin{figure*}[t]
\centering
\includegraphics[width=\textwidth]{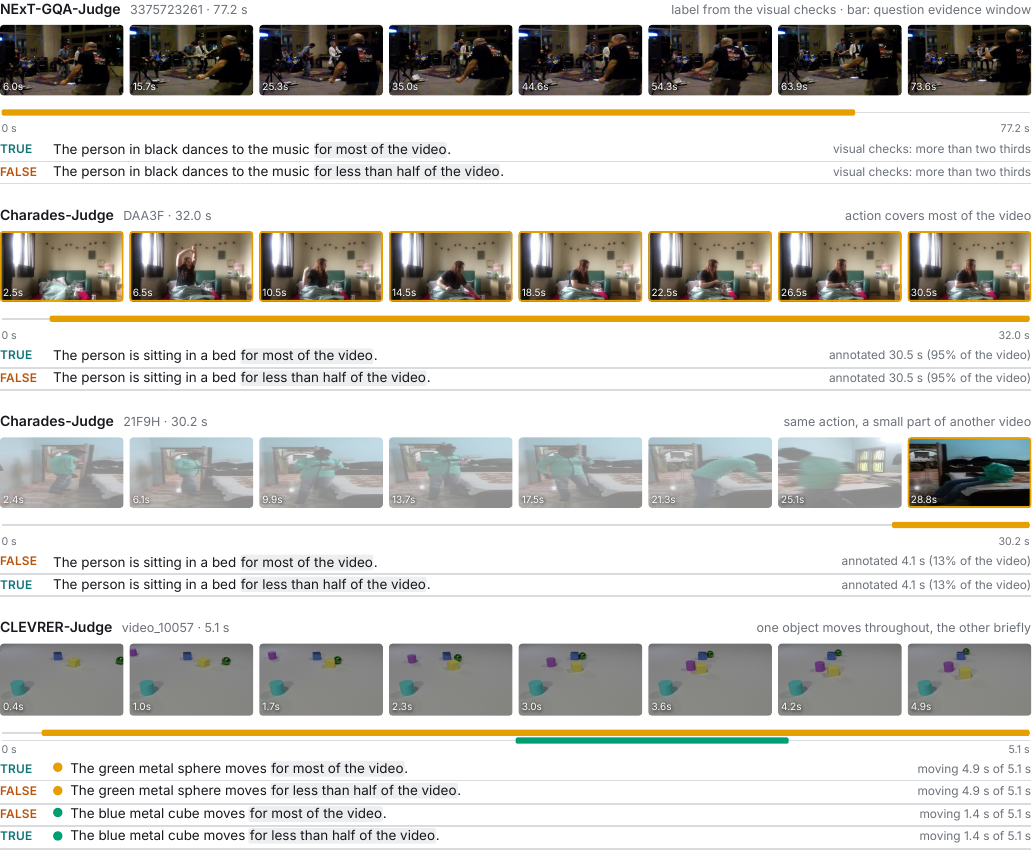}
\caption{
    \textbf{Duration claims.}
    Examples from the three judge sets.
    Positive and negative claims describe the same event but differ in whether it lasts for most or less than half of the video.
    Timelines show event durations for Ch-J and Cl-J, and question-evidence spans for NG-J.
}
\label{fig:Duration_claims}
\end{figure*}

%% file: figures/fig_C_frequency_claims.tex
\begin{figure*}[t]
\centering
\includegraphics[width=\textwidth]{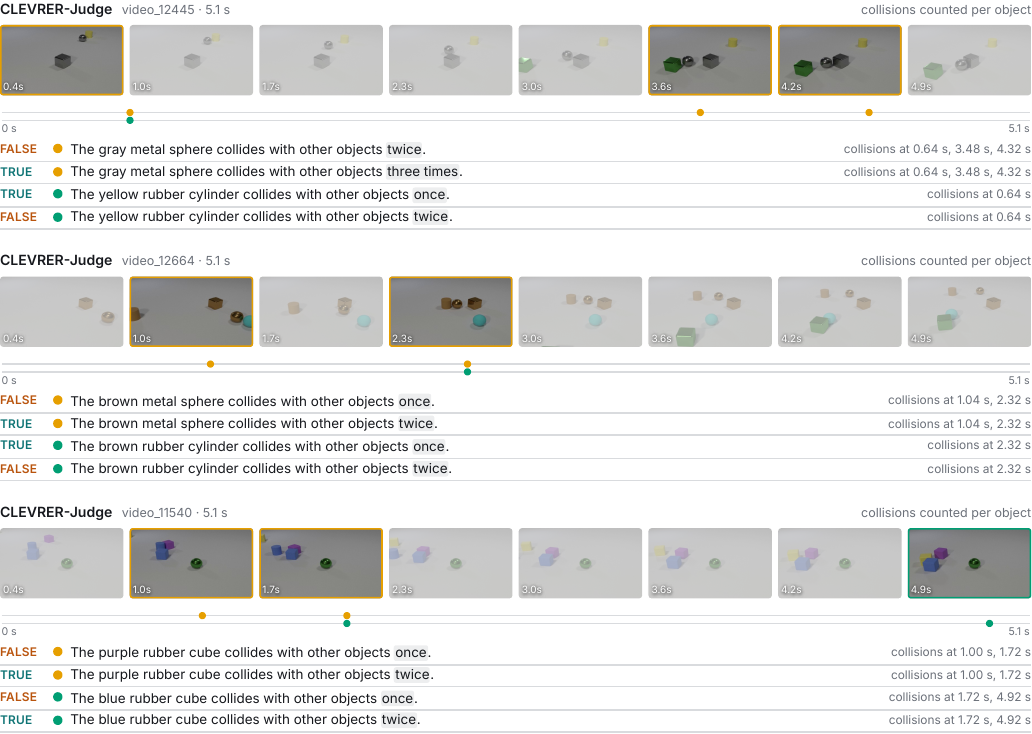}
\caption{
    \textbf{Frequency claims.}
    Examples from CLEVRER-Judge.
    Positive and negative claims describe the same object but differ in the number of collision events.
    Collision timestamps are indicated along the video timelines.
}
\label{fig:Frequency_claims}
\end{figure*}

%% file: sections/app_D_metrics.tex
\section{Metrics and Evaluation Protocol}
\label{app:Metrics_and_Evaluation_Protocol}

\paragraph{Pooled and within-video AUC.}
We evaluate judgment using pooled and within-video AUC, which measure complementary aspects of score discrimination.
Pooled AUC compares positive and negative claims across the entire evaluation set, whereas within-video AUC restricts comparisons to claims about the same video.
For order evaluation, let $s_i^{+}=s_\theta(v_i,c_i^{+})$ and $s_{i'}^{-}=s_\theta(v_{i'},c)$, where $c\in\mathcal{P}(c_{i'}^{+})\setminus\{c_{i'}^{+}\}$ is an order negative.
Both metrics use
\begin{equation*}
    \phi(s_i^{+},s_{i'}^{-})
    =\mathbf{1}[s_i^{+}>s_{i'}^{-}]
    +\tfrac12\,\mathbf{1}[s_i^{+}=s_{i'}^{-}].
\end{equation*}
Pooled AUC averages $\phi$ over all positive-negative pairs in the evaluation set.
Within-video order AUC averages $\phi(s_i^{+},s_i^{-})$ over the five order negatives of each video and then across videos.
Within-video content AUC similarly compares each chronological claim with its corresponding content negative, averaged over videos for which a valid negative is available.
The same AUC principle applies to localization, duration, and frequency claims.

\paragraph{Score comparability.}
For order evaluation, fewer than 0.4\% of pooled positive-negative pairs come from the same video, making pooled AUC sensitive to cross-video score variation.
In contrast, within-video comparisons cancel additive score offsets shared by claims from the same video.
\Cref{tab:Pooled_versus_within_video_evaluation} reports both metrics, the fraction of videos whose chronological claim ranks first among six candidates, and a diagnostic pooled AUC after subtracting each video's mean candidate score.
This diagnostic requires multiple candidate scores and is not available under independent single-claim inference.

\paragraph{Evaluation protocol.}
We compute each metric over eligible claims in the corresponding evaluation set (\Cref{tab:Judge_sets}), following the baseline-specific protocols in \Cref{app:Baseline_Adaptations_and_Inference_Cost}.
Paired analyses compare the same videos and candidates.
AUC assigns half credit to ties, whereas the task-format comparison in \Cref{sec:Diagnosing_Temporal_Judgment} requires a strictly positive score difference and counts ties as failures.

%% file: sections/app_E_notation.tex
\section{Notation}
\label{app:Notation}

\Cref{tab:Shared_notation} summarizes the notation used throughout the paper.
A parenthesized superscript indicates the training round, as in $\theta^{(r)}$.
A record $x_i$ represents one sampling instance of a source video, including its sampled segments and generated descriptions, from which training claims are subsequently constructed.
Multiple records may originate from the same source video.

\input{tables/tab_E_notation}

%% file: tables/tab_E_notation.tex
\begin{table}[ht]
\centering
\small
\setlength{\tabcolsep}{5pt}
\caption{
    \textbf{Notation summary.}
    Key symbols and functions used throughout the main paper and supplementary material.
}
\label{tab:Shared_notation}
\begin{tabular}{ll}
    \toprule
    Role & Notation\\
    \midrule
    Source / generated records & $\mathcal{D}_{\mathtt{source}}$, $\mathcal{D}_{\mathtt{generated}}$\\
    Training / evaluation data & $\mathcal{D}_{\mathtt{train}}^{(r)}$, $\mathcal{D}_{\mathtt{evaluation}}$\\
    Source-specific training data & $\mathcal{D}_{\mathtt{train},d}^{(r)}$\\
    Model / its score & $\pi_{\theta^{(r)}}$, $s_{\theta^{(r)}}$\\
    Video / segment / description & $v_i$, $u_{\{i,j\}}$, $e_{\{i,j\}}$\\
    True claim / its reorderings & $c_i^{+}$, $\mathcal{P}(c_i^{+})$\\
    Claim procedures & $\Template$, $\Swap$\\
    Order negatives / candidates & $\mathcal{N}_i$, $\mathcal{C}_i$\\
    Counts & $N_{\mathtt{source}}$, $N_{\mathtt{R}}$, $N_{\mathtt{G}}$, $N_{\mathtt{E}}$\\
    Learning objectives & $\mathcal{L}_{\mathtt{rank},i}$, $\mathcal{L}_{\mathtt{calibration},i}$\\
    Presence score & $a_{\{i,j,k\}}$\\
    Prompt labels & $q_{\mathtt{judge}}$, $q_{\mathtt{presence}}$\\
    Named functions & $\softplus$, $\BCE$, $\Unif$\\
    Construction procedures & $\Describe$, $\Select$, $\Train$\\
    Literal answer tokens & \tokYes{}, \tokNo{}\\
    \bottomrule
\end{tabular}
\end{table}

%% file: sections/app_F_data_construction.tex
\section{Data Construction and Semantic Validity}
\label{app:Data_Construction_and_Semantic_Validity}

The chronological order of sampled segments is known by construction, while generated event descriptions may contain inaccuracies or ambiguities.
We address these issues through confidence-based record selection and content-negative screening, which improve segment-description correspondence and reduce ambiguous training contrasts.

\paragraph{Record selection.}
We select training records based on their segment-description correspondence scores defined in \Cref{sec:Selecting_Segment_Grounded_Records}.
For each source dataset $d$ and sampling draw, the confidence threshold is determined by
\begin{equation}
\tau_d=Q_{1-\rho}\!\left(
\{\gamma_i:x_i\in\mathcal{D}_{\mathtt{generated},d}\}
\right),
\end{equation}
where $Q$ denotes the empirical quantile and $\rho=0.6$.
Records satisfying $\gamma_i\ge\tau_d$ are retained and aggregated across sources and draws to construct $\mathcal{D}_{\mathtt{train}}^{(1)}$.
Subsequent rounds follow the re-selection protocol described in \Cref{app:Additional_Rounds_of_Data_Selection}.

\paragraph{Content-negative screening.}
Content negatives are constructed by replacing the middle event description with one from another video of the same source dataset and sampling draw.
To reduce semantic overlap with the target video, we retain a content negative only when
\begin{equation}
\min_j a_{\{i,j,j\}}-\max_j s_{\theta^{(0)}}(u_{\{i,j\}},e_{\mathtt{foreign}};q_{\mathtt{presence}})\ge 2.
\end{equation}
This criterion requires the foreign event to receive lower presence scores than the original descriptions across the sampled segments.
Additional rounds update these scores using the preceding judge (\Cref{app:Additional_Rounds_of_Data_Selection}).

\paragraph{Training data.}
The source pool contains $N_{\mathtt{source}}=1{,}493$ videos, selected using source-dataset annotations without using those annotations as training targets.
We generate three sampling draws per video, filter invalid or repeated descriptions, and apply the confidence-based selection above.
This yields $N_{\mathtt{train}}^{(1)}=1{,}933$ training records for Qwen3-VL-8B.

%% file: sections/app_G_implementation.tex
\section{Implementation Details}
\label{app:Implementation_Details}

\subsection{Loss normalization}
\label{app:Loss_normalization}

The auxiliary objective in \Cref{eq:loss} combines calibration, content discrimination, and no-video regularization:
\begin{equation*}
\begin{split}
    \mathcal{L}_{\mathtt{aux},i}={}&\lambda_{\mathtt{calibration}}\mathcal{L}_{\mathtt{calibration},i}+\lambda_{\mathtt{content}}\mathcal{L}_{\mathtt{content},i}\\
    &+\lambda_{\mathtt{no\labelhyphen video}}\mathcal{L}_{\mathtt{no\labelhyphen video},i}.
\end{split}
\end{equation*}

Let $\BCE(s,y)$ denote binary cross-entropy with logits. The calibration loss supervises chronological claims as positive and reordered claims as negative:
\begin{equation}
\begin{split}
    \mathcal{L}_{\mathtt{calibration},i}={}&\BCE(s_\theta(v_i,c_i^{+}),1)\\
    &+\frac{1}{|\mathcal{N}_i|}\sum_{c\in\mathcal{N}_i}\BCE(s_\theta(v_i,c),0).
\end{split}
\end{equation}

The content loss additionally penalizes incorrect event descriptions when a valid content negative is available:
\begin{equation}
    \mathcal{L}_{\mathtt{content},i}=\BCE(s_\theta(v_i,c_{i,\mathtt{content}}^{-}),0).
\end{equation}

The no-video regularizer discourages preferences among candidate claims when visual evidence is absent. Given $\mathcal{C}_i=\{c_i^{+}\}\cup\mathcal{N}_i$, we define
\begin{equation}
    p_i(c)=\frac{\exp s_\theta(\varnothing,c)}
    {\sum_{c'\in\mathcal{C}_i}\exp s_\theta(\varnothing,c')},
\end{equation}
and minimize its divergence from the uniform distribution:
\begin{equation}
    \mathcal{L}_{\mathtt{no\labelhyphen video},i}
    =D_{\mathrm{KL}}\bigl(p_i\,\|\,\Unif(\mathcal{C}_i)\bigr).
\end{equation}
The KL divergence is computed from the model distribution to the uniform distribution over candidate claims, rather than answer tokens.

We use $m=2$, $\lambda_{\mathtt{calibration}}=\lambda_{\mathtt{content}}=0.2$, $\lambda_{\mathtt{no\labelhyphen video}}=1$, and $\lambda_{\mathtt{anchor}}=0.5$.
The content and anchor losses are set to zero when no valid content negative is available.

\subsection{Optimization}
\label{app:Optimization}

For Qwen3-VL, whole-video judgment uses 32 frames resized to $448\times256$, while segment description and scoring use 16 frames.
We apply rank-64 LoRA adapters ($\alpha=128$, dropout 0.05) to the attention and MLP projections of all decoder layers.
Training uses AdamW with a learning rate of $10^{-4}$, cosine scheduling, gradient accumulation over four records, and gradient clipping at norm 1.0.
Each model is trained for one epoch from the base checkpoint with a newly initialized adapter.
The content anchor uses the frozen base checkpoint without the adapter or gradient computation.

\paragraph{Backbones.}
Each backbone independently generates and selects training records using the same procedure, yielding 1,933 records for Qwen3-VL-8B, 1,549 for LLaVA-NeXT-Video-7B, and 2,424 for InternVL3.5-8B.
All three backbones use the same LoRA rank, adaptation of all decoder layers, and loss weights.
Answer-token IDs are determined separately for each tokenizer.

%% file: sections/app_H_prompts.tex
% Prompt box: a header strip with the name of the prompt (not part of the prompt) above the prompt text in monospace.
% \promptvar marks a slot that is filled per item.
% Inside a box, write "\ " after a period, question mark, or colon that is followed by a space; monospace would otherwise print two spaces.
\usetikzlibrary{backgrounds}
\newsavebox{\promptsavebox}
\newcommand{\promptvar}[1]{\textcolor{plotblue}{\char`\{#1\char`\}}}
\newcommand{\promptlabelfont}{\footnotesize\usefont{\encodingdefault}{\rmdefault}{m}{sc}\color{black!65}}
\makeatletter
\newenvironment{promptbox}[1]{%
  \def\promptboxlabel{#1}%
  \par\addvspace{0.5\baselineskip}\noindent
  \begin{lrbox}{\promptsavebox}%
  \begin{minipage}{\dimexpr\linewidth-16.4pt\relax}\footnotesize\fontfamily{cmtt}\selectfont\raggedright}
 {\end{minipage}\end{lrbox}%
  \begin{tikzpicture}[baseline=(b.base)]
    \node[anchor=north west,inner xsep=8pt,inner ysep=5.5pt,outer sep=0pt] (b) {\usebox{\promptsavebox}};
    \node[anchor=south west,inner xsep=8pt,inner ysep=2.4pt,outer sep=0pt,text width=\wd\promptsavebox,align=left] (t) at (b.north west) {\promptlabelfont\promptboxlabel};
    \begin{scope}[on background layer]
      \begin{scope}
        \clip[rounded corners=2.5pt] (t.north west) rectangle (b.south east);
        \fill[black!3] (t.north west) rectangle (b.south east);
        \fill[black!9] (t.north west) rectangle (t.south east);
      \end{scope}
      \draw[black!22,line width=0.4pt,rounded corners=2.5pt] (t.north west) rectangle (b.south east);
      \draw[black!22,line width=0.4pt] (t.south west) -- (t.south east);
    \end{scope}
  \end{tikzpicture}%
  \par\addvspace{0.5\baselineskip}\@endpetrue}
\makeatother

\section{Prompts}
\label{app:Prompts}

\paragraph{Judgment prompt.}
We use the following prompt for standalone video-claim judgment:
\begin{promptbox}{Judgment}
You are a strict video judge.\ You will be shown a video and a proposed description of it.\ Your only job is to decide whether the description is correct for this video.\\
Question:\ What happens in this video, and in what order?\\
Proposed answer:\ \promptvar{claim}\\
Is this answer correct given the video?\ Answer with only Yes or No.
\end{promptbox}
The diagnosis in \Cref{sec:Diagnosing_Temporal_Judgment} omits the opening sentence, ``You are a strict video judge.'', and uses the resulting diagnosis prompt.
Chronological claims follow the template ``First \{$e_1$\}. Then \{$e_2$\}. Finally \{$e_3$\}.''
Evaluation considers all six event permutations in the order $(1,2,3)$, $(2,1,3)$, $(3,2,1)$, $(1,3,2)$, $(2,3,1)$, and $(3,1,2)$, whereas training uses the three pairwise transpositions as order negatives.

\paragraph{Task formats of the diagnosis.}
The QA, Pair, and Single formats in \Cref{fig:Weak_Temporal_Order_Sensitivity_in_Judgment}(a) evaluate the same temporal fact, namely the order of two events $e_1$ and $e_2$.

\emph{QA} directly asks which event occurs first:
\begin{promptbox}{QA}
Which of these happens first in the video?\\
A. \promptvar{$e_1$}\\
B. \promptvar{$e_2$}\\
Answer with only A or B.
\end{promptbox}

\emph{Pair} compares two claims with opposite temporal orders:
\begin{promptbox}{Pair}
Here are two descriptions of the video.\\
A. \promptvar{claim}\\
B. \promptvar{claim with the events exchanged}\\
Which one is correct?\ Answer with only A or B.
\end{promptbox}

For QA and Pair, we compute the logit difference between the correct and incorrect options, averaged over both option positions.
\emph{Single} evaluates the two claims independently using the diagnosis prompt and compares their judgment scores.
A prediction is correct when the true alternative receives a strictly higher score.

\paragraph{Segment description prompt.}
We generate event descriptions from sampled video segments using:
\begin{promptbox}{Segment Description}
Describe the main visible action in this video clip as one short clause in the present tense, for example `a man opens a door'.\ Answer with only the clause.
\end{promptbox}
Descriptions are generated greedily with a maximum of 24 new tokens.

\paragraph{Segment presence prompt.}
We assess segment-description correspondence using:
\begin{promptbox}{Segment Presence}
Does the video visibly show this event:\ \promptvar{event}?\ Answer with only Yes or No.
\end{promptbox}
The correspondence score is the difference between the Yes and No logits under this presence prompt, rather than the whole-video judgment prompt.

%% file: sections/app_I_baselines.tex
\section{Baseline Adaptations and Inference Cost}
\label{app:Baseline_Adaptations_and_Inference_Cost}

This section details the evaluation protocols for the baselines in \Cref{tab:Temporal_judgment_and_content_discrimination}.
Throughout the paper, Qwen3-VL-8B refers to the instruction-tuned Qwen3-VL-8B-Instruct checkpoint~\cite{bai2025qwen3}.

\subsection{Released checkpoints}
\label{app:Released_checkpoints}

We evaluate released judges using their original backbones and official prompts, with the same 32 video frames and corresponding evaluation claims.
VideoJudge-7B~\cite{waheed2025videojudge} generates a rationale and a 1--5 rating, while LLaVA-Critic-7B~\cite{xiong2025llavacritic} and UnifiedReward-7B~\cite{wang2025unifiedreward} use their respective pointwise scoring protocols.
We additionally evaluate a TrustJudge-style expected-rating readout~\cite{wang2025trustjudge}, replacing the generated integer with its probability-weighted expectation.
Unparseable LLaVA-Critic ratings are excluded from AUC computation.
VideoJudge-7B and UnifiedReward-7B use Qwen2.5-VL-7B~\cite{bai2025qwen25vltechnicalreport}, while LLaVA-Critic-7B uses LLaVA-OneVision-7B~\cite{li2024llavaonevision}.
Because these checkpoints differ from our primary backbone, we report their absolute performance rather than changes from a shared base model.
A comparison on the Qwen2.5-VL-7B backbone is provided in \Cref{app:Qwen2_5_VL_backbone}.

\paragraph{Checkpoints on the Qwen3-VL-8B backbone.}
VideoSSR-8B~\cite{he2026videossr} uses the same backbone and is evaluated with our judgment prompt and score in \Cref{eq:score}.
UnifiedReward-Think-8B~\cite{wang2025unifiedrewardthink} supports pairwise video evaluation through its official prompt.
We evaluate each claim pair in both candidate positions and report pairwise accuracy (\%), averaged over the two positions.
These results are marked $\dagger$ in \Cref{tab:Temporal_judgment_and_content_discrimination} to distinguish them from pooled AUC.

\paragraph{Qwen3-VL-8B-Thinking.}
We also evaluate the released Qwen3-VL-8B-Thinking checkpoint without additional training.
With thinking enabled, the model generates up to 1,536 reasoning tokens before the judgment score is computed using \Cref{eq:score}.
Without thinking, the score is obtained directly from the first answer-token logits.
\Cref{tab:Reasoning_checkpoint_and_larger_backbone} compares both settings with the base judge and Ours.

\subsection{Training-free adaptations}
\label{app:Training_free_adaptations}

\paragraph{Distribution-sensitive scoring.}
We adapt the TrustJudge scoring approach to our backbone using a 1--5 rating prompt with a temporal-order rubric.
Correct event sequences receive high ratings, order errors receive intermediate ratings, and absent events receive the lowest rating.
The final score is the expected rating under the model's output distribution~\cite{liu2023g}.
This temporal-order rubric improves average order AUC from 69.3 with a generic rubric to 70.8, with improvements on all three judge sets.

\paragraph{Rationale and self-consistency.}
We also evaluate rationale generation~\cite{wei2022cot} and self-consistency~\cite{wang2023selfconsistency} as inference-time adaptations.
The rationale setting generates up to 256 tokens before producing a verdict, while self-consistency averages Yes-No logit differences over five sampled generations.
The single-rationale setting reduces average order AUC from 71.7 to 53.1, indicating that generating a rationale does not improve temporal judgment under this protocol.

\subsection{Inference cost}
\label{app:Inference_cost}

Our trained judge scores each video-claim pair with a single forward pass, whereas self-consistency requires five generations.
Claims about the same video share the video and instruction prefix (\Cref{app:Prompts}), so its key-value cache is computed once and reused across claims.
On 100 videos from the three judge sets, each with six order permutations, independent evaluation takes 0.52\,s per claim on an RTX A6000 GPU, or 3.14\,s per video.
Reusing the cached prefix reduces this to 0.78\,s per video, about 1.5 times the cost of a single claim, with the same order AUC of 79.8.
All results in this paper use independent forward passes.

%% file: sections/app_J_diagnostics.tex
\section{Extended Diagnostics}
\label{app:Extended_Diagnostics}

We conduct all diagnostics on 211 NG-J validation videos, disjoint from the test sets (\Cref{app:NExT_GQA_Judge}), using the diagnosis prompt (\Cref{app:Prompts}).
The task-format comparison uses two-event claims, while the linear probe and paired-score analyses use three-event claims.
Because the main results use the full judgment prompt, their absolute scores are not directly comparable to those obtained in the diagnostics.

\input{figures/fig_J_diagnosis_offsets}

\paragraph{Failures on facts answered correctly.}
We compare QA and Single on the same temporal fact across the 211 validation videos.
QA correctly identifies the event order in 135 videos, but Single succeeds on only 81 of these (60\%).
This discrepancy demonstrates that correct temporal understanding in QA does not consistently translate into correct standalone judgment, even for the same video.

\paragraph{Linear probe.}
We evaluate linear probes~\cite{alain2016probes} on frozen hidden states using out-of-fold predictions grouped by video.
The layer and regularization strength are selected within the training folds.
The probe achieves an AUC of 0.62, compared with 0.50 when trained on shuffled labels.
Chronological and reordered claims use the same event descriptions and template, controlling for differences in event content and vocabulary.
These results indicate that temporal order information is accessible in the hidden representations, but do not establish that the judgment scores use this information.

\input{tables/tab_K_backbone_q30}
\input{tables/tab_K_backbone_qwen25}

\paragraph{Score correlations.}
For three-event claims, chronological and reordered claim scores have a correlation of 0.89, compared with 0.43 between chronological claims and content negatives (\Cref{fig:Score_offsets_and_order_sensitivity}).
This stronger correlation is consistent with judgment scores being more sensitive to shared event content than to temporal order.

\paragraph{Removing the shared midpoint.}
Subtracting the shared midpoint $b_i$ from both scores of each pair increases pooled order AUC from 0.54 to 0.69.
This diagnostic illustrates how variation in shared score offsets can obscure temporal order discrimination across videos.
However, it requires both claims of each pair and cannot be applied to independent single-claim inference.

%% file: figures/fig_J_diagnosis_offsets.tex
\begin{figure}[tp]
\centering
\includegraphics[width=\linewidth]{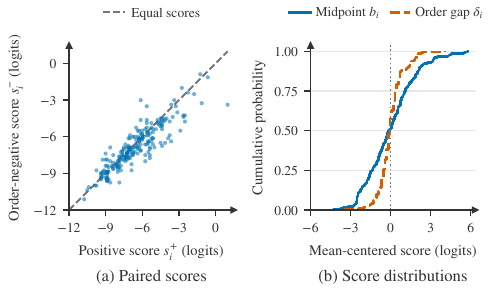}
\caption{
\textbf{Score offsets and order sensitivity.}
    Analysis of 211 NG-J validation videos.
    (a) Judgment scores of chronological and reordered claims, where points below the diagonal favor the chronological claim.
    (b) Empirical cumulative distributions of the shared midpoint $b_i$ and order gap $\delta_i$, each centered by its mean.
    Their standard deviations are 1.848 and 0.912 logits, respectively.
}
\label{fig:Score_offsets_and_order_sensitivity}
\end{figure}

%% file: tables/tab_K_backbone_q30.tex
\begin{table*}[t]
\centering
\footnotesize
\setlength{\tabcolsep}{2.3pt}
\caption{
    \textbf{Reasoning checkpoint and larger backbone.}
    Pooled test AUC $\times100$ using the metrics in \Cref{tab:Temporal_judgment_and_content_discrimination}.
    The upper block compares Qwen3-VL-8B-Thinking with and without reasoning against the base judge and Ours.
    The lower block evaluates Qwen3-VL-30B-A3B-Instruct, with LoRA applied only to attention projections due to its mixture-of-experts architecture.
    Parentheses indicate changes from the corresponding base model.
    \textbf{Bold} and \underline{underlined} values indicate the best and second-best results within each block.
}
\label{tab:Reasoning_checkpoint_and_larger_backbone}
\resizebox{\textwidth}{!}{%
\begin{tabular}{lccccccccccccccccc}
    \toprule
    & \multicolumn{4}{c}{NExT-GQA-Judge} & \multicolumn{4}{c}{Charades-Judge} & \multicolumn{5}{c}{CLEVRER-Judge} & \multicolumn{4}{c}{\textit{Average}}\\
    \cmidrule(lr){2-5}\cmidrule(lr){6-9}\cmidrule(lr){10-14}\cmidrule(lr){15-18}
    Method & Ord. & Loc. & Dur. & Content & Ord. & Loc. & Dur. & Content & Ord. & Loc. & Dur. & Freq. & Content & Ord. & Loc. & Dur. & Content\\
    \midrule
    Qwen3-VL-8B & 56.8 & 67.3 & 69.9 & \underline{84.2} & 71.6 & 87.7 & 71.0 & \underline{77.9} & 86.7 & 84.2 & 59.1 & 51.9 & 98.5 & 71.7 & 79.7 & 66.7 & \underline{86.9}\\
    Qwen3-VL-8B-Thinking~\cite{bai2025qwen3} & 55.5 & \underline{74.1} & \textbf{83.8} & 73.3 & 64.6 & \underline{90.6} & \textbf{88.5} & 72.0 & \textbf{95.1} & \textbf{98.1} & \textbf{79.6} & \textbf{78.5} & 99.6 & 71.7 & \underline{87.6} & \textbf{84.0} & 81.6\\
    Qwen3-VL-8B-Thinking, no thinking & \underline{59.1} & 71.2 & \underline{70.3} & 83.4 & \underline{74.7} & 90.0 & \underline{74.2} & 76.1 & \underline{93.3} & 87.0 & \underline{77.0} & \underline{59.6} & \textbf{100.0} & \underline{75.7} & 82.7 & \underline{73.8} & 86.5\\
    \rowcolor{oursrow}Qwen3-VL-8B + \textbf{Ours} & \textbf{62.9} & \textbf{83.5} & 66.9 & \textbf{91.7} & \textbf{81.8} & \textbf{95.6} & 69.3 & \textbf{85.7} & 92.4 & \underline{96.0} & 68.1 & 52.1 & \underline{99.7} & \textbf{79.0} & \textbf{91.7} & 68.1 & \textbf{92.4}\\
    \rowcolor{oursrow} & \dcolp{6.1} & \dcolp{16.2} & \dcolm{3.0} & \dcolp{7.5} & \dcolp{10.2} & \dcolp{7.9} & \dcolm{1.7} & \dcolp{7.8} & \dcolp{5.7} & \dcolp{11.8} & \dcolp{9.0} & \dcolp{0.2} & \dcolp{1.2} & \dcolp{7.3} & \dcolp{12.0} & \dcolp{1.4} & \dcolp{5.5}\\[1pt]
    \midrule
    Qwen3-VL-30B-A3B & 61.5 & 77.0 & 61.6 & 85.4 & 76.6 & 90.8 & 68.0 & 78.9 & 77.7 & 80.6 & 62.2 & 51.6 & 95.7 & 71.9 & 82.8 & 63.9 & 86.7\\
    \rowcolor{oursrow}Qwen3-VL-30B-A3B + \textbf{Ours} & \textbf{68.5} & \textbf{83.8} & \textbf{76.3} & \textbf{92.9} & \textbf{85.4} & \textbf{95.7} & \textbf{78.9} & \textbf{85.5} & \textbf{93.3} & \textbf{94.3} & \textbf{67.7} & \textbf{54.3} & \textbf{99.8} & \textbf{82.4} & \textbf{91.3} & \textbf{74.3} & \textbf{92.7}\\
    \rowcolor{oursrow} & \dcolp{7.0} & \dcolp{6.8} & \dcolp{14.7} & \dcolp{7.5} & \dcolp{8.8} & \dcolp{4.9} & \dcolp{10.9} & \dcolp{6.6} & \dcolp{15.6} & \dcolp{13.7} & \dcolp{5.5} & \dcolp{2.7} & \dcolp{4.1} & \dcolp{10.5} & \dcolp{8.5} & \dcolp{10.4} & \dcolp{6.0}\\[1pt]
    \bottomrule
\end{tabular}}
\end{table*}

%% file: tables/tab_K_backbone_qwen25.tex
\begin{table*}[t]
\centering
\footnotesize
\setlength{\tabcolsep}{2.6pt}
\caption{
    \textbf{Qwen2.5-VL-7B-Instruct backbone.}
    Pooled test AUC $\times100$ across the three judge sets.
    All methods use the same 32 video frames and evaluation candidates, while released judges retain their official prompts.
    The base judge, self-consistency, and Ours use the score in \Cref{eq:score}.
    Ours follows the Qwen3-VL training procedure, generating three sampling draws per video with this backbone.
}
\label{tab:Qwen2.5_VL_7B_Instruct_backbone}
\begin{tabular}{lr@{\hspace{\tabcolsep}}lr@{\hspace{\tabcolsep}}lr@{\hspace{\tabcolsep}}lr@{\hspace{\tabcolsep}}lr@{\hspace{\tabcolsep}}lr@{\hspace{\tabcolsep}}lr@{\hspace{\tabcolsep}}lr@{\hspace{\tabcolsep}}l}
    \toprule
    & \multicolumn{4}{c}{NExT-GQA-Judge} & \multicolumn{4}{c}{Charades-Judge} & \multicolumn{4}{c}{CLEVRER-Judge} & \multicolumn{4}{c}{\textit{Average}}\\
    \cmidrule(lr){2-5}\cmidrule(lr){6-9}\cmidrule(lr){10-13}\cmidrule(lr){14-17}
    Method & \multicolumn{2}{c}{Order} & \multicolumn{2}{c}{Content} & \multicolumn{2}{c}{Order} & \multicolumn{2}{c}{Content} & \multicolumn{2}{c}{Order} & \multicolumn{2}{c}{Content} & \multicolumn{2}{c}{Order} & \multicolumn{2}{c}{Content}\\
    \midrule
    \rowcolor{bandrow}\multicolumn{17}{c}{\texttt{Qwen2.5-VL-7B-Instruct}}\\
    \multicolumn{17}{l}{\textit{\textbf{Inference only}}}\\
    \quad base judge & 56.1 &  & 81.5 &  & 67.0 &  & 70.6 &  & 79.8 &  & 99.9 &  & 67.6 &  & 84.0 & \\
    \quad base judge + TrustJudge~\cite{wang2025trustjudge} & 54.7 & \dcolm{1.4} & 77.5 & \dcolm{4.0} & 59.4 & \dcolm{7.6} & 71.5 & \dcolp{0.9} & 73.5 & \dcolm{6.3} & 99.8 & \dcolm{0.1} & 62.5 & \dcolm{5.1} & 83.0 & \dcolm{1.0}\\
    \quad base judge + self-consistency~\cite{wang2023selfconsistency} & 56.3 & \dcolp{0.2} & 72.2 & \dcolm{9.3} & 67.5 & \dcolp{0.5} & 71.0 & \dcolp{0.4} & 72.8 & \dcolm{7.0} & 99.9 & \dcolz & 65.5 & \dcolm{2.1} & 81.0 & \dcolm{3.0}\\
    \cmidrule(l{2pt}r{2pt}){1-17}
    \multicolumn{17}{l}{\textit{\textbf{Trained by prior work} (released checkpoints on this backbone)}}\\
    \quad UnifiedReward-7B~\cite{wang2025unifiedreward} & 51.4 & \dcolm{4.7} & 56.9 & \dcolm{24.6} & 54.1 & \dcolm{12.9} & 58.2 & \dcolm{12.4} & 66.7 & \dcolm{13.1} & 96.2 & \dcolm{3.7} & 57.4 & \dcolm{10.2} & 70.4 & \dcolm{13.6}\\
    \quad UnifiedReward-7B + TrustJudge~\cite{wang2025trustjudge} & 52.7 & \dcolm{3.4} & 75.7 & \dcolm{5.8} & 58.1 & \dcolm{8.9} & 72.1 & \dcolp{1.5} & 74.3 & \dcolm{5.5} & 99.5 & \dcolm{0.4} & 61.7 & \dcolm{5.9} & 82.4 & \dcolm{1.6}\\
    \quad VideoJudge-7B~\cite{waheed2025videojudge} & 50.5 & \dcolm{5.6} & 56.2 & \dcolm{25.3} & 50.4 & \dcolm{16.6} & 51.2 & \dcolm{19.4} & 50.0 & \dcolm{29.8} & 50.0 & \dcolm{49.9} & 50.3 & \dcolm{17.3} & 52.5 & \dcolm{31.5}\\
    \quad VideoJudge-7B + TrustJudge~\cite{wang2025trustjudge} & 52.0 & \dcolm{4.1} & 62.3 & \dcolm{19.2} & 52.1 & \dcolm{14.9} & 54.6 & \dcolm{16.0} & 54.1 & \dcolm{25.7} & 82.2 & \dcolm{17.7} & 52.7 & \dcolm{14.9} & 66.3 & \dcolm{17.7}\\
    \cmidrule(l{2pt}r{2pt}){1-17}
    \multicolumn{17}{l}{\textit{\textbf{Trained on self-generated comparisons}}}\\
    \rowcolor{oursrow}\quad \textbf{Ours} & \textbf{66.3} & \dcolp{10.2} & \textbf{90.0} & \dcolp{8.5} & \textbf{80.1} & \dcolp{13.1} & \textbf{76.3} & \dcolp{5.7} & \textbf{97.2} & \dcolp{17.4} & \textbf{100.0} & \dcolp{0.1} & \textbf{81.2} & \dcolp{13.6} & \textbf{88.8} & \dcolp{4.8}\\
    \bottomrule
\end{tabular}
\end{table*}

%% file: sections/app_K_backbones.tex
\section{Additional Backbones}
\label{app:Additional_Backbones}

We extend the experiments in \Cref{sec:Main_Results} to two additional backbones, evaluating the generality of our framework across model scales and architectures.

\subsection{Larger backbone}
\label{app:Larger_backbone}

We apply our framework to Qwen3-VL-30B-A3B-Instruct, adapting the attention projections with LoRA.
As shown in the lower block of \Cref{tab:Reasoning_checkpoint_and_larger_backbone}, Ours improves all reported metrics, including duration across all three judge sets.
Average AUC increases by 10.5, 8.5, 10.4, and 6.0 points for order, localization, duration, and content, respectively.

\subsection{Qwen2.5-VL backbone}
\label{app:Qwen2_5_VL_backbone}

We further evaluate Qwen2.5-VL-7B-Instruct, the backbone used by UnifiedReward-7B and VideoJudge-7B.
As shown in \Cref{tab:Qwen2.5_VL_7B_Instruct_backbone}, Ours improves order AUC by 10.2, 13.1, and 17.4 points on NG-J, Ch-J, and Cl-J, respectively, while also improving content AUC on all three sets.
In contrast, self-consistency and the released judges do not improve average order AUC over the base model.

%% file: sections/app_L_ablations.tex
\input{figures/fig_L_additional_rounds}

\input{tables/tab_M_within_video}

\section{Additional Ablations}
\label{app:Additional_Ablations}

\subsection{Ablation protocols}
\label{app:Ablation_protocols}

In \Cref{tab:Ablations}, the variants without the content anchor or no-video regularization set the corresponding loss weights to zero.
The random-label variant assigns a uniformly sampled target ordering to each training record.
All three variants use the same 1,933 records and random seed as Ours.
The second-round variant reselects training records using the updated judge, as described below.

\subsection{Additional rounds of data selection}
\label{app:Additional_Rounds_of_Data_Selection}

We investigate whether iterative updates to the judge improve subsequent training rounds (\Cref{sec:Iterating_the_Self_Improving_Loop}).
At each subsequent round, the preceding judge reselects the same number of training records, and a new judge is initialized from the base checkpoint and trained with the same objective.
We compare two generator settings: reusing descriptions from the base model and regenerating descriptions with the preceding judge.

\Cref{fig:Additional_rounds} shows that neither generator achieves consistent improvements beyond the first round.
Order, localization, and content performance remain relatively stable over eight rounds, with average order AUC ranging from 76.8 to 81.6 and content AUC from 90.6 to 92.4.
Duration and frequency fluctuate across rounds without a consistent trend.
We therefore use the first round for the main results.

%% file: figures/fig_L_additional_rounds.tex
\begin{figure*}[t]
    \centering
    \includegraphics[width=\textwidth]{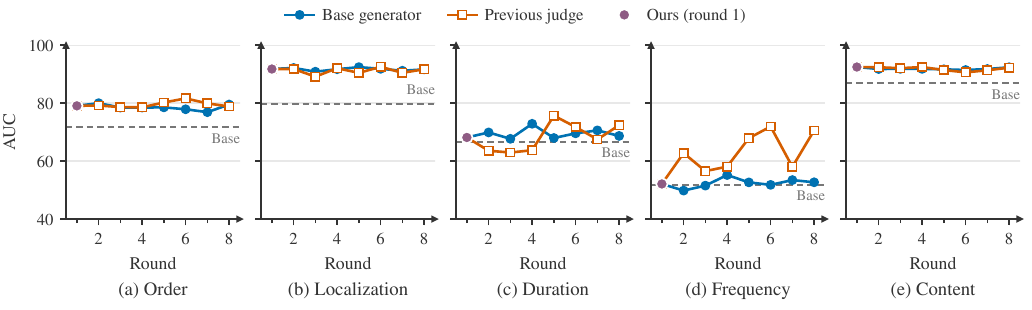}
    \caption{
        \textbf{Additional rounds.}
        Average pooled test AUC $\times100$ across training rounds with two generator settings on Qwen3-VL-8B.
        Metrics follow \Cref{tab:Temporal_judgment_and_content_discrimination}.
        Both settings coincide in the first round (Ours), and dashed lines indicate the base model.
    }
    \label{fig:Additional_rounds}
\end{figure*}

%% file: tables/tab_M_within_video.tex
\begin{table*}[t]
\centering
\footnotesize
\setlength{\tabcolsep}{4.2pt}
\caption{
    \textbf{Pooled versus within-video evaluation.}
    Test AUC $\times100$ for Qwen3-VL-8B (Base) and Ours.
    Pooled AUC compares positive and negative claims across the evaluation set, whereas within-video AUC restricts comparisons to claims from the same video.
    Ranked first indicates the percentage of videos whose chronological claim scores highest among six candidates (chance: 16.7\%).
    $^{*}$Pooled AUC after subtracting each video's mean candidate score, requiring all six candidates rather than independent single-claim inference.
    Parentheses indicate changes from Base.
}
\label{tab:Pooled_versus_within_video_evaluation}
\begin{tabular}{lr@{\hspace{\tabcolsep}}lr@{\hspace{\tabcolsep}}lr@{\hspace{\tabcolsep}}lr@{\hspace{\tabcolsep}}lr@{\hspace{\tabcolsep}}lr@{\hspace{\tabcolsep}}lr@{\hspace{\tabcolsep}}lr@{\hspace{\tabcolsep}}l}
\toprule
 & \multicolumn{4}{c}{NExT-GQA-Judge} & \multicolumn{4}{c}{Charades-Judge} & \multicolumn{4}{c}{CLEVRER-Judge} & \multicolumn{4}{c}{\textit{Average}}\\
\cmidrule(lr){2-5}\cmidrule(lr){6-9}\cmidrule(lr){10-13}\cmidrule(lr){14-17}
Evaluation & \multicolumn{2}{c}{Order} & \multicolumn{2}{c}{Content} & \multicolumn{2}{c}{Order} & \multicolumn{2}{c}{Content} & \multicolumn{2}{c}{Order} & \multicolumn{2}{c}{Content} & \multicolumn{2}{c}{Order} & \multicolumn{2}{c}{Content}\\
\midrule
\multicolumn{17}{l}{\textit{\textbf{Pooled AUC} (\Cref{tab:Temporal_judgment_and_content_discrimination})}}\\
\quad Base & 56.8 &  & 84.2 &  & 71.6 &  & 77.9 &  & 86.7 &  & 98.5 &  & 71.7 &  & 86.9 & \\
\rowcolor{oursrow}\quad \textbf{Ours} & 62.9 & \dcolp{6.1} & 91.7 & \dcolp{7.5} & 81.8 & \dcolp{10.2} & 85.7 & \dcolp{7.8} & 92.4 & \dcolp{5.7} & 99.7 & \dcolp{1.2} & 79.0 & \dcolp{7.3} & 92.4 & \dcolp{5.5}\\
\cmidrule(l{2pt}r{2pt}){1-17}
\multicolumn{17}{l}{\textit{\textbf{Within-video AUC}}}\\
\quad Base & 64.0 &  & 93.3 &  & 81.9 &  & 88.6 &  & 93.0 &  & 99.7 &  & 79.7 &  & 93.9 & \\
\rowcolor{oursrow}\quad \textbf{Ours} & 67.4 & \dcolp{3.4} & 94.8 & \dcolp{1.5} & 84.1 & \dcolp{2.2} & 88.0 & \dcolm{0.6} & 93.2 & \dcolp{0.2} & 100.0 & \dcolp{0.3} & 81.6 & \dcolp{1.9} & 94.2 & \dcolp{0.3}\\
\cmidrule(l{2pt}r{2pt}){1-17}
\multicolumn{17}{l}{\textit{\textbf{True claim ranked first of six}}}\\
\quad Base & 27.2 &  & -- &  & 54.4 &  & -- &  & 74.5 &  & -- &  & 52.0 &  & -- & \\
\rowcolor{oursrow}\quad \textbf{Ours} & 37.4 & \dcolp{10.2} & -- &  & 59.7 & \dcolp{5.3} & -- &  & 79.0 & \dcolp{4.5} & -- &  & 58.7 & \dcolp{6.7} & -- & \\
\cmidrule(l{2pt}r{2pt}){1-17}
\multicolumn{17}{l}{\textit{\textbf{Pooled AUC, per-video offsets removed}$^{*}$}}\\
\quad Base & 66.4 &  & 92.9 &  & 82.9 &  & 85.6 &  & 91.6 &  & 99.1 &  & 80.3 &  & 92.5 & \\
\rowcolor{oursrow}\quad \textbf{Ours} & 71.1 & \dcolp{4.7} & 94.0 & \dcolp{1.1} & 87.1 & \dcolp{4.2} & 87.0 & \dcolp{1.4} & 94.7 & \dcolp{3.1} & 99.8 & \dcolp{0.7} & 84.3 & \dcolp{4.0} & 93.6 & \dcolp{1.1}\\
\bottomrule
\end{tabular}
\end{table*}

%% file: sections/app_M_analysis.tex
\section{Analysis of Judgment Improvements}
\label{app:Analysis_of_Judgment_Improvements}

\subsection{Pooled and within-video evaluation}
\label{app:Pooled_and_within_video_evaluation}

\Cref{tab:Pooled_versus_within_video_evaluation} compares pooled and within-video AUC for the base judge and Ours on the test sets, using the metrics defined in \Cref{app:Metrics_and_Evaluation_Protocol}.
Average order AUC improves by 7.3 points in pooled evaluation and 1.9 points in within-video evaluation.
The larger pooled gain indicates improved discrimination across videos, beyond the gains observed within individual videos.

\paragraph{Comparison at inference.}
We additionally evaluate explicit comparisons among alternative event orderings at inference.
Scoring all six permutations of a three-event claim and subtracting the highest competing score increases average order AUC from 71.7 to 81.4 for the base judge and from 79.0 to 84.9 for Ours.
This procedure requires six forward passes and applies only when the claim's events can be reordered, whereas our method scores each claim independently.
The additional improvement suggests that standalone temporal judgment has further room for improvement.

\subsection{Score decomposition after training}
\label{app:Score_decomposition_after_training}

Across all three test sets, training increases both the mean order gap $\delta_i$ and the spread of shared score offsets $b_i$.
The mean gap grows relative to the offset standard deviation, consistent with the validation analysis in \Cref{fig:Score_decomposition_before_and_after_training}.
Thus, stronger order preferences accompany the improved discrimination even though cross-video offset variation persists.

\subsection{Frame-order controls}
\label{app:Frame_order_controls}

\Cref{tab:Frame_order_controls} evaluates the test claims with shuffled and reversed video frames across all three backbones.
For every trained judge, shuffling moves order AUC toward chance, while reversing the frames reduces it below chance.
These findings are consistent with sensitivity to the visual order of events.
Content AUC is less affected overall, as changing frame order does not alter whether an unrelated event is present.

\input{tables/tab_M_frame_controls}

\subsection{QA and judgment after training}
\label{app:QA_and_judgment_after_training}

\Cref{tab:QA_and_judgment_before_and_after_training} compares QA and independent judgment before and after training on the NG-J validation and test sets, extending the diagnostics in \Cref{app:Extended_Diagnostics}.
Judgment uses the main judgment prompt.
For two-event claims, within-video judgment accuracy increases from 47.9\% to 61.6\% on validation and from 57.7\% to 62.0\% on test, approaching QA performance without improvements in QA accuracy.
For three-event claims, pooled AUC improves on both splits, but test within-video accuracy changes little (58.4\% to 59.3\%), and cases where QA succeeds but judgment fails increase from 59 to 77.
Thus, improved score discrimination does not uniformly resolve the QA--judgment discrepancy.

\input{tables/tab_M_matched_after_training}

%% file: tables/tab_M_frame_controls.tex
\begin{table}[t]
\centering
\footnotesize
\setlength{\tabcolsep}{5pt}
\caption{
    \textbf{Frame-order controls.}
    Pooled test AUC $\times100$ for each backbone before and after training, evaluated with original, randomly shuffled, and reversed video frames.
    Content evaluation on Ch-J uses 299 videos with valid content negatives.
}
\label{tab:Frame_order_controls}
\resizebox{\columnwidth}{!}{%
    \begin{tabular}{llcccccc}
    \toprule
    & & \multicolumn{2}{c}{NG-J} & \multicolumn{2}{c}{Ch-J} & \multicolumn{2}{c}{Cl-J}\\
    \cmidrule(lr){3-4}\cmidrule(lr){5-6}\cmidrule(lr){7-8}
    Model & Frames & Ord. & Content & Ord. & Content & Ord. & Content\\
    \midrule
    Qwen3-VL-8B & Original & 56.8 & 84.2 & 71.6 & 77.9 & 86.7 & 98.5\\
    & Shuffled & 52.4 & 83.1 & 54.1 & 73.3 & 54.9 & 99.2\\
    & Reversed & 49.0 & 83.0 & 40.0 & 69.3 & 36.4 & 92.4\\
    \midrule
    \rowcolor{oursrow}\quad + \textbf{Ours} & Original & 62.9 & 91.7 & 81.8 & 85.7 & 92.4 & 99.7\\
    \rowcolor{oursrow} & Shuffled & 51.5 & 90.7 & 53.0 & 78.9 & 46.5 & 99.7\\
    \rowcolor{oursrow} & Reversed & 45.1 & 91.2 & 39.0 & 75.6 & 33.8 & 98.8\\
    \midrule
    LLaVA-NeXT-Video-7B & Original & 52.4 & 70.4 & 52.0 & 62.8 & 48.9 & 55.7\\
    & Shuffled & 50.7 & 70.4 & 51.4 & 62.5 & 48.9 & 55.7\\
    & Reversed & 51.9 & 71.0 & 50.5 & 61.7 & 48.5 & 54.3\\
    \midrule
    \rowcolor{oursrow}\quad + \textbf{Ours} & Original & 67.4 & 72.5 & 78.6 & 64.4 & 53.9 & 53.7\\
    \rowcolor{oursrow} & Shuffled & 54.1 & 75.3 & 50.7 & 63.7 & 49.2 & 59.8\\
    \rowcolor{oursrow} & Reversed & 43.9 & 65.9 & 33.5 & 55.0 & 42.2 & 51.4\\
    \midrule
    InternVL3.5-8B & Original & 58.3 & 83.1 & 74.2 & 78.5 & 92.5 & 100.0\\
    & Shuffled & 51.3 & 82.4 & 51.9 & 75.7 & 50.1 & 99.6\\
    & Reversed & 46.1 & 82.9 & 37.5 & 70.8 & 22.8 & 97.8\\
    \midrule
    \rowcolor{oursrow}\quad + \textbf{Ours} & Original & 66.4 & 89.0 & 86.4 & 83.0 & 99.0 & 99.9\\
    \rowcolor{oursrow} & Shuffled & 52.0 & 88.8 & 50.2 & 78.5 & 47.2 & 99.9\\
    \rowcolor{oursrow} & Reversed & 41.9 & 87.5 & 33.6 & 70.5 & 20.6 & 99.2\\
    \bottomrule
    \end{tabular}%
}
\end{table}

%% file: tables/tab_M_matched_after_training.tex
\begin{table}[t]
\centering
\footnotesize
\setlength{\tabcolsep}{3pt}
\caption{
    \textbf{QA and judgment before and after training.}
    Comparison of Qwen3-VL-8B (Base) and Ours on NG-J.
    QA evaluates the order of the first two events, while Single and Three-event denote judgment of two- and three-event claims using the main prompt.
    QA and Within report accuracy (\%, with ties counted as failures), and Pooled reports AUC $\times100$.
    Three-event rows compare each chronological claim with the reordering that swaps its first two events.
    QA only counts videos where QA succeeds but judgment fails.
}
\label{tab:QA_and_judgment_before_and_after_training}
\resizebox{\columnwidth}{!}{%
    \begin{tabular}{llrrrrrrrr}
    \toprule
    & & \multicolumn{2}{c}{QA} & \multicolumn{2}{c}{Within} & \multicolumn{2}{c}{Pooled} & \multicolumn{2}{c}{QA only}\\
    \cmidrule(lr){3-4}\cmidrule(lr){5-6}\cmidrule(lr){7-8}\cmidrule(lr){9-10}
    Videos & Claim & Base & Ours & Base & Ours & Base & Ours & Base & Ours\\
    \midrule
    Val. (211) & Single & 64.0 & 59.7 & 47.9 & 61.6 & 51.1 & 57.2 & 56 & 21\\
    & Three-event & 64.0 & 59.7 & 56.4 & 68.7 & 53.9 & 60.0 & 45 & 38\\
    \midrule
    Test (305) & Single & 63.0 & 63.0 & 57.7 & 62.0 & 54.4 & 58.8 & 58 & 39\\
    & Three-event & 63.0 & 63.0 & 58.4 & 59.3 & 54.2 & 58.9 & 59 & 77\\
    \bottomrule
    \end{tabular}%
}
\end{table}

%% file: sections/app_N_transfer.tex
\section{Transfer to Other Temporal Claims}
\label{app:Transfer_to_Other_Temporal_Claims}

\subsection{Event count and template}
\label{app:Event_count_and_template}

\Cref{tab:Event_count_and_claim_template} evaluates claims containing two to four annotated events across three sentence templates.
Although Ours is trained only on three-event claims using a single template, improvements in both order and content AUC persist across the evaluated event counts when averaged over the three templates.

\input{tables/tab_N_event_format}

\input{tables/tab_N_relations}

\subsection{Explicit temporal relations}
\label{app:Explicit_temporal_relations}

We construct temporal relation claims from annotated action intervals in Ch-J test videos using the template ``The person is \{action\} \{connective\} \{action\}.''
For sequential actions separated by a clear temporal gap, true claims use \emph{before} or \emph{after} in the correct direction, while false claims reverse the direction or use \emph{while}.
For actions overlapping for most of the shorter action, true claims use \emph{while}, while false claims use \emph{before} or \emph{after}.
The evaluation includes 322 videos with sequential action pairs, of which 255 also contain overlapping pairs.

\input{tables/tab_N_temporal_main}

\Cref{tab:Transfer_to_temporal_relations} shows that Ours improves discrimination between correct and reversed \emph{before}/\emph{after} relations in both pooled and within-video evaluation.
It also better distinguishes sequential actions from incorrect \emph{while} claims, although identifying overlapping actions remains challenging for both models.
These results demonstrate transfer to explicit temporal order relations, with more limited improvements for overlap.

\subsection{Localization, duration, and frequency}
\label{app:Localization_duration_and_frequency_results}

\Cref{app:Temporal_attribute_claims} describes the construction of these evaluation sets, while \Cref{tab:Temporal_judgment_and_content_discrimination} reports their pooled AUC.
For Qwen3-VL-8B, Ours improves localization across all three sets and duration on Cl-J, while duration on NG-J and Ch-J does not improve and collision counting remains near chance.
The larger Qwen3-VL-30B-A3B backbone improves duration across all three sets (\Cref{app:Larger_backbone}).
\Cref{tab:Temporal_attribute_claims} provides additional comparisons between QA and judgment, along with two control settings.

\paragraph{QA versus judgment.}
We compare two formats for evaluating the same temporal fact: a question with two answer options (QA) and two independently scored claims (Judge).
Both formats are evaluated by whether the correct alternative receives a higher score.
Ties receive half credit.
For duration, the base model performs approximately 9 to 13 percentage points better in QA than in judgment, whereas localization shows smaller differences.
This discrepancy suggests that the difference between answering and judging extends beyond event order.
After training, within-event judgment accuracy improves for both localization and duration across all three sets, even when pooled duration AUC does not improve.

\paragraph{Controls.}
We examine sensitivity to question wording and visual input.
Replacing ``What happens in this video, and in what order?'' with ``What happens in this video?'' preserves localization improvements across all three sets.
Without video input, pooled AUC generally remains near chance for both models, consistent with the balanced temporal expressions.
These results support the role of visual evidence in localization improvements, even without an explicit temporal order question.

%% file: tables/tab_N_event_format.tex
\begin{table}[t]
\centering
\footnotesize

\setlength{\tabcolsep}{2.4pt}
\caption{
    \textbf{Event count and claim template.}
    Pooled test AUC $\times100$ for different event counts, averaged over three sentence templates.
    Base judge columns report absolute AUC, while other columns report changes from Base.
    Both trained variants use only three-event claims with one template, and w/o anchor excludes the content anchor.
    Four-event evaluation uses 58 Ch-J videos with four temporally separated events.
}
\label{tab:Event_count_and_claim_template}
\begin{tabular}{lcrcccccc}
\toprule
    & & & \multicolumn{2}{c}{Base judge} & \multicolumn{2}{c}{Ours} & \multicolumn{2}{c}{w/o anchor}\\
    \cmidrule(lr){4-5}\cmidrule(lr){6-7}\cmidrule(lr){8-9}
    Set & $N$ & Videos & Order & Content & Order & Content & Order & Content\\
    \midrule
    NG-J & 2 & 305 & 56.4 & 89.1 & $+4.2$ & $+1.9$ & $+10.6$ & $-2.6$\\
    & 3 & 305 & 57.1 & 85.0 & $+5.8$ & $+6.4$ & $+13.4$ & $+2.0$\\
    \midrule
    Ch-J & 2 & 330 & 71.8 & 82.2 & $+7.2$ & $+1.5$ & $+9.7$ & $-6.2$\\
    & 3 & 330 & 70.9 & 78.6 & $+10.2$ & $+6.2$ & $+13.8$ & $-1.8$\\
    & 4 & 58 & 72.0 & 75.4 & $+10.7$ & $+7.0$ & $+14.6$ & $-2.8$\\
    \midrule
    Cl-J & 2 & 331 & 80.4 & 97.7 & $+9.8$ & $+2.0$ & $+14.7$ & $+0.8$\\
    & 3 & 331 & 82.4 & 95.6 & $+9.9$ & $+4.1$ & $+12.4$ & $+2.8$\\
\bottomrule
\end{tabular}
\end{table}

%% file: tables/tab_N_relations.tex
\begin{table}[t]
\centering
\footnotesize
\setlength{\tabcolsep}{4pt}
\caption{
    \textbf{Transfer to temporal relations.}
    Pooled and within-video AUC $\times100$ on Ch-J test videos with sequential (322 videos) or overlapping (255 videos) action pairs.
    The three comparisons evaluate reversed \emph{before}/\emph{after} relations, sequential actions against false \emph{while} claims, and overlapping actions against false \emph{before}/\emph{after} claims.
}
\label{tab:Transfer_to_temporal_relations}
\begin{tabular}{lcccc}
    \toprule
    & \multicolumn{2}{c}{Pooled} & \multicolumn{2}{c}{Within video}\\
    \cmidrule(lr){2-3}\cmidrule(lr){4-5}
    True claim against false claim & Base & Ours & Base & Ours\\
    \midrule
    Correct against reversed order & 63.6 & \textbf{73.1} & 74.5 & \textbf{82.8}\\
    Sequential pair against \emph{while} & 60.0 & \textbf{68.9} & 76.1 & \textbf{86.1}\\
    Overlapping pair against \emph{before} or \emph{after} & 45.9 & \textbf{48.6} & 34.4 & \textbf{44.9}\\
    \bottomrule
\end{tabular}
\end{table}

%% file: tables/tab_N_temporal_main.tex
\begin{table*}[t]
\centering
\footnotesize
\setlength{\tabcolsep}{5pt}
\caption{
    \textbf{QA, judgment, and temporal controls.}
    Left: Within-event accuracy (\%) comparing two-option QA with independently scored claims.
    Right: Pooled AUC $\times100$ under the original judgment prompt (\Cref{tab:Temporal_judgment_and_content_discrimination}), a neutral question, and no video input.
}
\label{tab:Temporal_attribute_claims}
\begin{tabular}{llrcccccccccc}
    \toprule
    & & & \multicolumn{4}{c}{Accuracy within each event} & \multicolumn{6}{c}{Pooled AUC}\\
    \cmidrule(lr){4-7}\cmidrule(lr){8-13}
    & & & \multicolumn{2}{c}{Base} & \multicolumn{2}{c}{Ours} & \multicolumn{2}{c}{Judgment prompt} & \multicolumn{2}{c}{Neutral question} & \multicolumn{2}{c}{Without video}\\
    \cmidrule(lr){4-5}\cmidrule(lr){6-7}\cmidrule(lr){8-9}\cmidrule(lr){10-11}\cmidrule(lr){12-13}
    Type & Set & Events & QA & Judge & QA & Judge & Base & Ours & Base & Ours & Base & Ours\\
    \midrule
    Loc. & NG-J & 96 & 78.6 & 75.5 & 90.6 & 90.6 & 67.3 & 83.5 & 69.1 & 82.6 & 49.9 & 53.8\\
    Loc. & Ch-J & 296 & 91.4 & 92.2 & 94.1 & 94.3 & 87.7 & 95.6 & 89.2 & 95.2 & 53.3 & 52.1\\
    Loc. & Cl-J & 478 & 91.9 & 91.6 & 95.1 & 96.4 & 84.2 & 96.0 & 86.5 & 96.0 & 49.6 & 50.7\\
    \midrule
    Dur. & NG-J & 64 & 93.8 & 81.2 & 90.6 & 88.3 & 69.9 & 66.9 & 71.6 & 67.5 & 53.8 & 57.1\\
    Dur. & Ch-J & 178 & 91.9 & 82.3 & 91.6 & 87.9 & 71.0 & 69.3 & 71.0 & 68.6 & 50.0 & 50.0\\
    Dur. & Cl-J & 658 & 76.6 & 67.8 & 70.3 & 76.4 & 59.1 & 68.1 & 64.1 & 67.6 & 50.3 & 49.6\\
    \midrule
    Freq. & Cl-J & 990 & 57.4 & 50.2 & 57.1 & 50.8 & 51.9 & 52.1 & 52.3 & 52.0 & 52.2 & 49.8\\
    \bottomrule
\end{tabular}
\end{table*}

%% file: sections/app_O_external.tex
\input{tables/tab_O_videoqa}

\section{External Benchmarks and Video QA}
\label{app:External_Benchmarks_Video_QA_and_Structural_Transfer}

\subsection{External judgment benchmarks}
\label{app:External_judgment_benchmarks}

\input{figures/fig_P_additional_corrected}

\input{figures/fig_P_qualitative}

\Cref{tab:External_judgment_benchmarks} reports pooled AUC using independently computed judgment scores on three human-verified benchmarks, measuring discrimination between statements that match the video and those that do not.

For TempCompass~\cite{liu2024tempcompass}, we evaluate yes/no questions using the difference between Yes and No logits.
\emph{Order} includes 382 event-order questions, while \emph{Others} combines the remaining four dimensions (2,071 questions).
For VELOCITI~\cite{saravanan2025velociti}, we use its official entailment prompt with 32 frames.
\emph{Order} covers 1,234 event chronology pairs, while \emph{Others} combines six additional tests, excluding the control test.
For Vinoground~\cite{zhang2024vinoground}, we use the same entailment prompt to compare each video with its own caption and the paired video's caption, which describes the same events in a different order.
The official Vinoground text, video, and group scores improve from 40.8, 47.6, and 32.6 to 58.4, 63.8, and 49.4, respectively.

\subsection{Video QA}
\label{app:Video_QA}

\Cref{tab:General_video_QA_after_judge_training} evaluates general video QA performance on NExT-QA~\cite{xiao2021nextqa} and TempCompass~\cite{liu2024tempcompass} without additional QA training.
Accuracy is largely preserved, with changes below one point on NExT-QA and the caption matching and yes/no tasks of TempCompass.
TempCompass multiple-choice accuracy decreases by approximately two points.
The content anchor reduces the performance drop on NExT-QA.

%% file: tables/tab_O_videoqa.tex
\begin{table}[t]
\centering
\small
\setlength{\tabcolsep}{3.6pt}
\caption{
    \textbf{General video QA after judge training.}
    Accuracy (\%) without additional QA training on NExT-QA (5,173 multiple-choice questions outside the judge construction pool) and TempCompass (1,580 multiple-choice, 1,503 caption matching, and 2,453 yes/no questions).
    Ours includes the content anchor, unlike the variant without it.
    Parentheses indicate changes from the base judge.
}
\label{tab:General_video_QA_after_judge_training}
\begin{tabular}{lcccc}
    \toprule
    & NExT-QA & \multicolumn{3}{c}{TempCompass}\\
    \cmidrule(lr){2-2}\cmidrule(lr){3-5}
    Model & MC & MC & Caption & Yes/No\\
    \midrule
    Base judge & 82.39 & 74.11 & 82.37 & 75.87\\
    w/o content anchor & \vdc{80.30}{\dminus{2.09}} & \vdc{72.47}{\dminus{1.65}} & \vdc{82.63}{\dplus{0.27}} & \vdc{77.99}{\dplus{2.12}}\\
    \rowcolor{oursrow}\textbf{Ours} & \vdc{81.67}{\dminus{0.72}} & \vdc{71.96}{\dminus{2.15}} & \vdc{82.37}{\dns{+0.00}} & \vdc{75.83}{\dminus{0.04}}\\
    \bottomrule
\end{tabular}
\end{table}

%% file: figures/fig_P_additional_corrected.tex
\begin{figure*}[t]
\centering
\includegraphics[width=\textwidth]{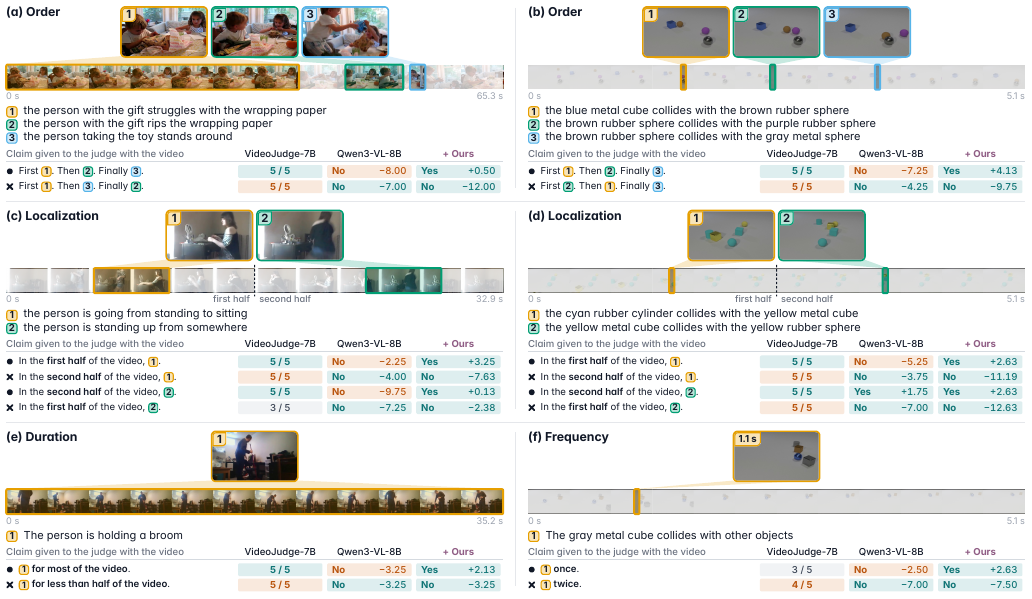}
\caption{\textbf{Additional corrected judgments.} Six further test videos in the format of \Cref{fig:Selected_corrected_judgments}. (a, b) Event order on NG-J and Cl-J. (c, d) Localization on Ch-J and Cl-J. (e) Duration on Ch-J. (f) Frequency on Cl-J. In every panel the base judge rejects at least one true claim, whereas Ours accepts every true claim and rejects every false one.}
\label{fig:Additional_corrected_judgments}
\end{figure*}

%% file: figures/fig_P_qualitative.tex
\begin{figure*}[t]
\centering
\includegraphics[width=\textwidth]{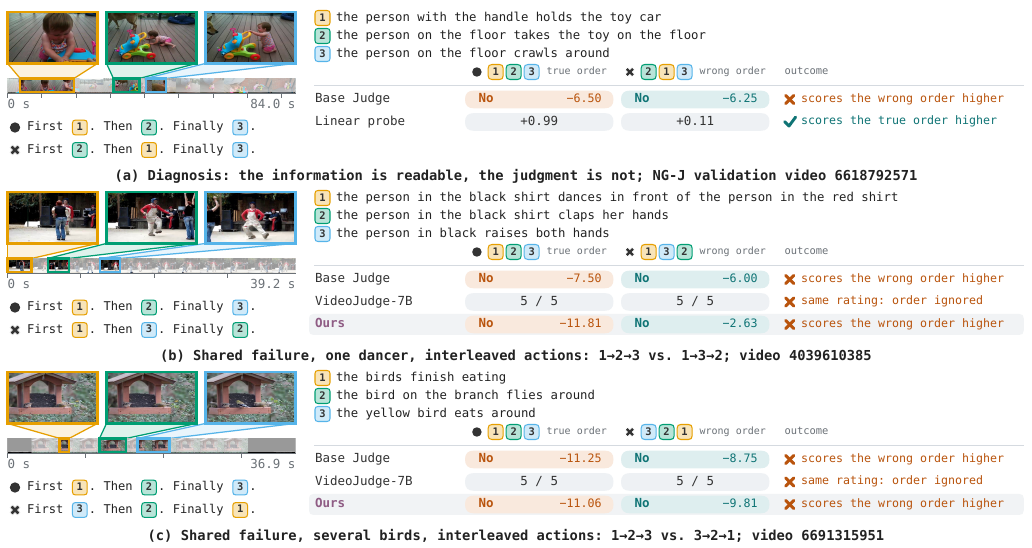}
\caption{\textbf{Diagnosis case and shared failures.} (a) On a validation video under the diagnosis prompt, the base judge answers \tokNo{} to both claims with nearly equal scores, while a supervised linear probe on its hidden states ranks them correctly. (b, c) All three judges fail on near-duplicate or interleaved actions. Cells follow \Cref{fig:Selected_corrected_judgments}; the probe gives a score, not a verdict.}
\label{fig:Diagnosis_case_and_shared_failures}
\end{figure*}

%% file: sections/app_P_qualitative.tex
\section{Qualitative Examples}
\label{app:Qualitative_Examples}

\Cref{fig:Selected_corrected_judgments}(a) illustrates an order judgment where the base judge accepts both the chronological and reordered claims, while Ours accepts only the chronological claim.
Panels (b) to (d) show selected localization, duration, and frequency examples where Ours correctly distinguishes true from false claims (\Cref{app:Temporal_attribute_claims}).
Aggregate results are reported in \Cref{tab:Temporal_judgment_and_content_discrimination}.

\Cref{fig:Additional_corrected_judgments} shows six further selected corrections across all three judge sets, where Ours accepts true claims rejected by the base judge while rejecting their false counterparts.
The duration and frequency cases illustrate individual corrections within the limited aggregate transfer reported in \Cref{app:Localization_duration_and_frequency_results}.

\Cref{fig:Diagnosis_case_and_shared_failures} presents an additional diagnostic example and two failures shared by all evaluated judges.
The failures involve similar or interleaved actions, such as closely related movements by a dancer or overlapping activities among birds, making temporal order difficult to distinguish from the sampled frames.

%% file: main.bib
@inproceedings{waheed2025videojudge,
  title = {{VideoJudge: Bootstrapping Enables Scalable Supervision of MLLM-as-a-Judge for Video Understanding}},
  author = {Abdul Waheed and Zhen Wu and Dareen Alharthi and Seungone Kim and Bhiksha Raj},
  booktitle = {International Conference on Learning Representations},
  year = {2026},
  url = {https://arxiv.org/abs/2509.21451}
}

@article{lin2025selfjudges,
  title={{Self-Improving VLM Judges Without Human Annotations}},
  author={Lin, Inna Wanyin and Hu, Yushi and Li, Shuyue Stella and Geng, Scott and Koh, Pang Wei and Zettlemoyer, Luke and Althoff, Tim and Ghazvininejad, Marjan},
  journal={arXiv preprint arXiv:2512.05145},
  year={2025},
  url = {https://arxiv.org/abs/2512.05145}
}

@article{huang2026evovid,
  title = {{EvoVid: Temporal-Centric Self-Evolution for Video Large Language Models}},
  author = {Shiqi Huang and Ziyue Wang and Zhongrong Zuo and Han Qiu and Qi She and Bihan Wen},
  journal = {arXiv preprint arXiv:2605.21931},
  year = {2026},
  url = {https://arxiv.org/abs/2605.21931}
}

@inproceedings{xiao2024nextgqa,
  title = {{Can I Trust Your Answer? Visually Grounded Video Question Answering}},
  author = {Junbin Xiao and Angela Yao and Yicong Li and Tat Seng Chua},
  booktitle = {Proceedings of the IEEE/CVF Conference on Computer Vision and Pattern Recognition},
  year = {2024},
  url = {https://arxiv.org/abs/2309.01327}
}

@inproceedings{sigurdsson2016charades,
  title = {{Hollywood in Homes: Crowdsourcing Data Collection for Activity Understanding}},
  author = {Gunnar A. Sigurdsson and G{\"u}l Varol and Xiaolong Wang and Ali Farhadi and Ivan Laptev and Abhinav Gupta},
  booktitle = {Proceedings of the European Conference on Computer Vision},
  pages = {510--526},
  year = {2016},
  url = {https://arxiv.org/abs/1604.01753}
}

@inproceedings{yi2020clevrer,
  title = {{CLEVRER: CoLlision Events for Video REpresentation and Reasoning}},
  author = {Kexin Yi and Chuang Gan and Yunzhu Li and Pushmeet Kohli and Jiajun Wu and Antonio Torralba and Joshua B. Tenenbaum},
  booktitle = {International Conference on Learning Representations},
  year = {2020},
  url = {https://arxiv.org/abs/1910.01442}
}

@inproceedings{hu2021lora,
  title = {{LoRA: Low-Rank Adaptation of Large Language Models}},
  author = {Edward J. Hu and Yelong Shen and Phillip Wallis and Zeyuan Allen-Zhu and Yuanzhi Li and Shean Wang and Lu Wang and Weizhu Chen},
  booktitle = {International Conference on Learning Representations},
  year = {2022},
  url = {https://arxiv.org/abs/2106.09685}
}

@inproceedings{liu2024tempcompass,
  title = {{TempCompass: Do Video LLMs Really Understand Videos?}},
  author = {Yuanxin Liu and Shicheng Li and Yi Liu and Yuxiang Wang and Shuhuai Ren and Lei Li and Sishuo Chen and Xu Sun and Lu Hou},
  booktitle = {Findings of the Association for Computational Linguistics: ACL 2024},
  pages = {8731--8772},
  year = {2024},
  url = {https://aclanthology.org/2024.findings-acl.517/}
}

@inproceedings{xiao2021nextqa,
  title = {{NExT-QA: Next Phase of Question-Answering to Explaining Temporal Actions}},
  author = {Junbin Xiao and Xindi Shang and Angela Yao and Tat-Seng Chua},
  booktitle = {Proceedings of the IEEE/CVF Conference on Computer Vision and Pattern Recognition},
  year = {2021},
  url = {https://arxiv.org/abs/2105.08276}
}

@article{wang2025trustjudge,
  title = {{TrustJudge: Inconsistencies of LLM-as-a-Judge and How to Alleviate Them}},
  author = {Yidong Wang and Yunze Song and Tingyuan Zhu and Xuanwang Zhang and Zhuohao Yu and Hao Chen and Chiyu Song and Qiufeng Wang and Cunxiang Wang and Zhen Wu and Xinyu Dai and Yue Zhang and Wei Ye and Shikun Zhang},
  journal = {arXiv preprint arXiv:2509.21117},
  year = {2025},
  url = {https://arxiv.org/abs/2509.21117}
}

@inproceedings{wang2023selfconsistency,
  title = {{Self-Consistency Improves Chain of Thought Reasoning in Language Models}},
  author = {Xuezhi Wang and Jason Wei and Dale Schuurmans and Quoc Le and Ed Chi and Sharan Narang and Aakanksha Chowdhery and Denny Zhou},
  booktitle = {International Conference on Learning Representations},
  year = {2023},
  url = {https://arxiv.org/abs/2203.11171}
}

@inproceedings{hewitt2019probes,
  title = {{Designing and Interpreting Probes with Control Tasks}},
  author = {John Hewitt and Percy Liang},
  booktitle = {Proceedings of EMNLP-IJCNLP},
  pages = {2733--2743},
  year = {2019},
  url = {https://aclanthology.org/D19-1275/}
}

@inproceedings{wei2022cot,
  title={Chain-of-Thought Prompting Elicits Reasoning in Large Language Models},
  author={Wei, Jason and Wang, Xuezhi and Schuurmans, Dale and Bosma, Maarten and Ichter, Brian and Xia, Fei and Chi, Ed and Le, Quoc and Zhou, Denny},
  booktitle={Advances in Neural Information Processing Systems},
  year={2022},
  url = {https://arxiv.org/abs/2201.11903}
}

@inproceedings{xiong2025llavacritic,
  title={{LLaVA-Critic}: Learning to Evaluate Multimodal Models},
  author={Xiong, Tianyi and Wang, Xiyao and Guo, Dong and Ye, Qinghao and Fan, Haoqi and Gu, Quanquan and Huang, Heng and Li, Chunyuan},
  booktitle={Proceedings of the IEEE/CVF Conference on Computer Vision and Pattern Recognition},
  year={2025},
  url = {https://openaccess.thecvf.com/content/CVPR2025/html/Xiong_LLaVA-Critic_Learning_to_Evaluate_Multimodal_Models_CVPR_2025_paper.html},
  pages = {13618--13628}
}

@article{wang2025unifiedreward,
  title={Unified Reward Model for Multimodal Understanding and Generation},
  author={Wang, Yibin and Zang, Yuhang and Li, Hao and Jin, Cheng and Wang, Jiaqi},
  journal={arXiv preprint arXiv:2503.05236},
  year={2025},
  url = {https://arxiv.org/abs/2503.05236}
}

@article{li2026unsupervisedmrm,
  title = {{Unsupervised Multimodal Reward Modeling}},
  author = {Yueying Li and Jingxuan Fan and Kiant{\'e} Brantley and Sham M. Kakade and Hanlin Zhang},
  journal = {arXiv preprint arXiv:2610.12707},
  year = {2026},
  url = {https://arxiv.org/abs/2610.12707}
}

@article{ianaro2026ordermatters,
  title = {{Order Matters: LVLMs as Judges for Temporal Reasoning in Image Sequences}},
  author = {Martina Ianaro and Guilherme Fernandes and Maurizio Gabbrielli and Joao Magalhaes},
  journal = {arXiv preprint arXiv:2608.10908},
  year = {2026},
  url = {https://arxiv.org/abs/2608.10908}
}

@inproceedings{sadhu2027verifiable,
  title = {{What Do Verifiable Rewards Teach Video-Language Models About Time? A Controlled Multi-Model Study}},
  author = {Avyay Sadhu and Patrick Cooper},
  booktitle = {Proceedings of the IEEE/CVF Winter Conference on Applications of Computer Vision},
  year = {2027},
  url = {https://arxiv.org/abs/2610.03792}
}

@article{li2026isovrbench,
  title = {{IsoVRBench: Video Reward Modeling Without Human Feedback}},
  author = {Shuyue Stella Li and Jacqueline He and Jiahui Chen and Sreya Dutta Roy and Yulia Tsvetkov and Luke Zettlemoyer and Marjan Ghazvininejad and Xiaochuang Han},
  journal = {arXiv preprint arXiv:2610.17234},
  year = {2026},
  url = {https://arxiv.org/abs/2610.17234}
}

@inproceedings{he2026videossr,
  title = {{VideoSSR: Video Self-Supervised Reinforcement Learning}},
  author = {Zefeng He and Xiaoye Qu and Yafu Li and Siyuan Huang and Daizong Liu and Yu Cheng},
  booktitle = {Proceedings of the IEEE/CVF Conference on Computer Vision and Pattern Recognition},
  year = {2026},
  url = {https://arxiv.org/abs/2511.06281}
}

@article{zhang2026videozero,
  title = {{Video-Zero: Self-Evolution Video Understanding}},
  author = {Ruixu Zhang and Deyi Ji and Lanyun Zhu and Xuanyi Liu and Yuxin Meng and Ruihang Chu and Yujiu Yang},
  journal = {arXiv preprint arXiv:2605.14733},
  year = {2026},
  url = {https://arxiv.org/abs/2605.14733}
}

@inproceedings{wang2025unifiedrewardthink,
  title = {{Unified Multimodal Chain-of-Thought Reward Model through Reinforcement Fine-Tuning}},
  author = {Yibin Wang and Zhimin Li and Yuhang Zang and Chunyu Wang and Qinglin Lu and Cheng Jin and Jiaqi Wang},
  booktitle = {Advances in Neural Information Processing Systems},
  year = {2025},
  url = {https://arxiv.org/abs/2505.03318}
}

@inproceedings{li2026temple,
  title = {{TEMPLE: Incentivizing Temporal Understanding of Video Large Language Models via Progressive Pre-SFT Alignment}},
  author = {Shicheng Li and Lei Li and Kun Ouyang and Shuhuai Ren and Yuanxin Liu and Yuanxing Zhang and Fuzheng Zhang and Lingpeng Kong and Qi Liu and Xu Sun},
  booktitle = {Proceedings of the AAAI Conference on Artificial Intelligence},
  year = {2026},
  url = {https://arxiv.org/abs/2503.16929}
}

@article{li2025tpo,
  title = {{Temporal Preference Optimization for Long-Form Video Understanding}},
  author = {Rui Li and Xiaohan Wang and Yuhui Zhang and Orr Zohar and Zeyu Wang and Serena Yeung-Levy},
  journal = {arXiv preprint arXiv:2501.13919},
  year = {2025},
  url = {https://arxiv.org/abs/2501.13919}
}

@inproceedings{kulkarni2025videopasta,
  title = {{VideoPASTA: 7K Preference Pairs That Matter for Video-LLM Alignment}},
  author = {Yogesh Kulkarni and Pooyan Fazli},
  booktitle = {Proceedings of the Conference on Empirical Methods in Natural Language Processing},
  year = {2025},
  url = {https://arxiv.org/abs/2504.14096}
}

@inproceedings{feng2025videor1,
  title = {{Video-R1: Reinforcing Video Reasoning in MLLMs}},
  author = {Kaituo Feng and Kaixiong Gong and Bohao Li and Zonghao Guo and Yibing Wang and Tianshuo Peng and Junfei Wu and Xiaoying Zhang and Benyou Wang and Xiangyu Yue},
  booktitle = {Advances in Neural Information Processing Systems},
  year = {2025},
  url = {https://arxiv.org/abs/2503.21776}
}

@inproceedings{xue2025arrowoftime,
  title = {{Seeing the Arrow of Time in Large Multimodal Models}},
  author = {Zihui Xue and Mi Luo and Kristen Grauman},
  booktitle = {Advances in Neural Information Processing Systems},
  year = {2025},
  url = {https://arxiv.org/abs/2506.03340}
}

@article{wang2024selftaught,
  title = {{Self-Taught Evaluators}},
  author = {Tianlu Wang and Ilia Kulikov and Olga Golovneva and Ping Yu and Weizhe Yuan and Jane Dwivedi-Yu and Richard Yuanzhe Pang and Maryam Fazel-Zarandi and Jason Weston and Xian Li},
  journal = {arXiv preprint arXiv:2408.02666},
  year = {2024},
  url = {https://arxiv.org/abs/2408.02666}
}

@article{nvidia2025cosmos,
  title = {{Cosmos} World Foundation Model Platform for Physical {AI}},
  author = {{NVIDIA}},
  journal = {arXiv preprint arXiv:2501.03575},
  year = {2025},
  url = {https://arxiv.org/abs/2501.03575}
}

@article{zheng2025vbench2,
  title = {{VBench-2.0}: Advancing Video Generation Benchmark Suite for Intrinsic Faithfulness},
  author = {Dian Zheng and Ziqi Huang and Hongbo Liu and Kai Zou and Yinan He and Fan Zhang and Lulu Gu and Yuanhan Zhang and Jingwen He and Wei-Shi Zheng and Yu Qiao and Ziwei Liu},
  journal = {arXiv preprint arXiv:2503.21755},
  year = {2025},
  url = {https://arxiv.org/abs/2503.21755}
}

@inproceedings{park2026perceptionjudge,
  title = {{Mitigating Perceptual Judgment Bias in Multimodal LLM-as-a-Judge via Perceptual Perturbation and Reward Modeling}},
  author = {Seojeong Park and Jiho Choi and Junyong Kang and Seonho Lee and Jaeyo Shin and Hyunjung Shim},
  booktitle = {International Conference on Machine Learning},
  year = {2026},
  url = {https://arxiv.org/abs/2606.02578}
}

@article{wang2025internvl35,
  title={{InternVL3.5: Advancing Open-Source Multimodal Models in Versatility, Reasoning, and Efficiency}},
  author={Wang, Weiyun and Gao, Zhangwei and Gu, Lixin and Pu, Hengjun and Cui, Long and Wei, Xingguang and Liu, Zhaoyang and Jing, Linglin and Ye, Shenglong and Shao, Jie and others},
  journal={arXiv preprint arXiv:2508.18265},
  year={2025},
  url = {https://arxiv.org/abs/2508.18265}
}

@article{yang2025qwen3,
  title={{Qwen3 Technical Report}},
  author={Yang, An and Li, Anfeng and Yang, Baosong and Zhang, Beichen and Hui, Binyuan and Zheng, Bo and Yu, Bowen and Gao, Chang and Huang, Chengen and Lv, Chenxu and others},
  journal={arXiv preprint arXiv:2505.09388},
  year={2025},
  url = {https://arxiv.org/abs/2505.09388}
}

@inproceedings{chen2024mllmjudge,
  title={{MLLM-as-a-Judge: Assessing Multimodal LLM-as-a-Judge with Vision-Language Benchmark}},
  author={Chen, Dongping and Chen, Ruoxi and Zhang, Shilin and Wang, Yaochen and Liu, Yinuo and Zhou, Huichi and Zhang, Qihui and Wan, Yao and Zhou, Pan and Sun, Lichao},
  booktitle={International Conference on Machine Learning},
  year={2024},
  url = {https://proceedings.mlr.press/v235/chen24h.html}
}

@inproceedings{pu2025judgeanything,
  title={{Judge Anything: MLLM as a Judge Across Any Modality}},
  author={Pu, Shu and Wang, Yaochen and Chen, Dongping and Chen, Yuhang and Wang, Guohao and Qin, Qi and Zhang, Zhongyi and Zhang, Zhiyuan and Zhou, Zetong and Gong, Shuang and others},
  booktitle={Proceedings of the ACM SIGKDD Conference on Knowledge Discovery and Data Mining},
  pages={5742--5753},
  year={2025},
  url = {https://arxiv.org/abs/2503.17489}
}

@inproceedings{lee2024prometheusvision,
  title={{Prometheus-Vision: Vision-Language Model as a Judge for Fine-Grained Evaluation}},
  author={Lee, Seongyun and Kim, Seungone and Park, Sue and Kim, Geewook and Seo, Minjoon},
  booktitle={Findings of the Association for Computational Linguistics: ACL 2024},
  pages={11286--11315},
  year={2024},
  url = {https://aclanthology.org/2024.findings-acl.672/}
}

@inproceedings{ko2025flexjudge,
  title={{Flex-Judge: Text-Only Reasoning Unleashes Zero-Shot Multimodal Evaluators}},
  author={Ko, Jongwoo and Kim, Sungnyun and Cho, Sungwoo and Yun, Se-Young},
  booktitle={Advances in Neural Information Processing Systems},
  year={2025},
  url = {https://arxiv.org/abs/2505.18601}
}

@inproceedings{kim2024prometheus2,
  title={{Prometheus 2: An Open Source Language Model Specialized in Evaluating Other Language Models}},
  author={Kim, Seungone and Suk, Juyoung and Longpre, Shayne and Lin, Bill Yuchen and Shin, Jamin and Welleck, Sean and Neubig, Graham and Lee, Moontae and Lee, Kyungjae and Seo, Minjoon},
  booktitle={Proceedings of the Conference on Empirical Methods in Natural Language Processing},
  year={2024},
  url = {https://aclanthology.org/2024.emnlp-main.248/}
}

@inproceedings{zhu2025judgelm,
  title={{JudgeLM: Fine-tuned Large Language Models are Scalable Judges}},
  author={Zhu, Lianghui and Wang, Xinggang and Wang, Xinlong},
  booktitle={International Conference on Learning Representations},
  year={2025},
  url = {https://arxiv.org/abs/2310.17631}
}

@inproceedings{li2025generationjudgment,
  title={{From Generation to Judgment: Opportunities and Challenges of LLM-as-a-Judge}},
  author={Li, Dawei and Jiang, Bohan and Huang, Liangjie and Beigi, Alimohammad and Zhao, Chengshuai and Tan, Zhen and Bhattacharjee, Amrita and Jiang, Yuxuan and Chen, Canyu and Wu, Tianhao and others},
  booktitle={Proceedings of the Conference on Empirical Methods in Natural Language Processing},
  pages={2757--2791},
  year={2025},
  url = {https://aclanthology.org/2025.emnlp-main.138/}
}

@inproceedings{yuan2024selfrewarding,
  title={{Self-Rewarding Language Models}},
  author={Yuan, Weizhe and Pang, Richard Yuanzhe and Cho, Kyunghyun and Li, Xian and Sukhbaatar, Sainbayar and Xu, Jing and Weston, Jason E},
  booktitle={International Conference on Machine Learning},
  year={2024},
  url = {https://arxiv.org/abs/2401.10020}
}

@inproceedings{jiang2024hacl,
  title={{Hallucination Augmented Contrastive Learning for Multimodal Large Language Model}},
  author={Jiang, Chaoya and Xu, Haiyang and Dong, Mengfan and Chen, Jiaxing and Ye, Wei and Yan, Ming and Ye, Qinghao and Zhang, Ji and Huang, Fei and Zhang, Shikun},
  booktitle={Proceedings of the IEEE/CVF Conference on Computer Vision and Pattern Recognition},
  pages={27036--27046},
  year={2024},
  url = {https://openaccess.thecvf.com/content/CVPR2024/html/Jiang_Hallucination_Augmented_Contrastive_Learning_for_Multimodal_Large_Language_Model_CVPR_2024_paper.html}
}

@inproceedings{park2026momentmix,
  title={{MomentMix Augmentation with Length-Aware DETR for Temporally Robust Moment Retrieval}},
  author={Park, Seojeong and Choi, Jiho and Baek, Kyungjune and Shim, Hyunjung},
  booktitle={Proceedings of the IEEE/CVF Winter Conference on Applications of Computer Vision},
  pages={1127--1136},
  year={2026},
  url = {https://openaccess.thecvf.com/content/WACV2026/html/Park_MomentMix_Augmentation_with_Length-Aware_DETR_for_Temporally_Robust_Moment_Retrieval_WACV_2026_paper.html}
}

@article{bai2025qwen3,
  title={{Qwen3-VL Technical Report}},
  author={Bai, Shuai and Cai, Yuxuan and Chen, Ruizhe and Chen, Keqin and Chen, Xionghui and Cheng, Zesen and Deng, Lianghao and Ding, Wei and Gao, Chang and Ge, Chunjiang and others},
  journal={arXiv preprint arXiv:2511.21631},
  year={2025},
  url = {https://arxiv.org/abs/2511.21631}
}

@article{bai2025qwen25vltechnicalreport,
  title={{Qwen2.5-VL Technical Report}},
  author={Bai, Shuai and Chen, Keqin and Liu, Xuejing and Wang, Jialin and Ge, Wenbin and Song, Sibo and Dang, Kai and Wang, Peng and Wang, Shijie and Tang, Jun and others},
  journal={arXiv preprint arXiv:2502.13923},
  year={2025},
  url = {https://arxiv.org/abs/2502.13923}
}

@article{li2024llavaonevision,
  title={{LLaVA-OneVision: Easy Visual Task Transfer}},
  author={Li, Bo and Zhang, Yuanhan and Guo, Dong and Zhang, Renrui and Li, Feng and Zhang, Hao and Zhang, Kaichen and Zhang, Peiyuan and Li, Yanwei and Liu, Ziwei and others},
  journal={arXiv preprint arXiv:2408.03326},
  year={2024},
  url = {https://arxiv.org/abs/2408.03326}
}

@misc{zhang2024llavanextvideo,
  title={{LLaVA-NeXT: A Strong Zero-shot Video Understanding Model}},
  author={Zhang, Yuanhan and Li, Bo and Liu, Haotian and Lee, Yong Jae and Gui, Liangke and Fu, Di and Feng, Jiashi and Liu, Ziwei and Li, Chunyuan},
  howpublished={\url{https://llava-vl.github.io/blog/2024-04-30-llava-next-video/}},
  month={April},
  year={2024}
}

@inproceedings{zheng2023judging,
  title={{Judging LLM-as-a-Judge with MT-Bench and Chatbot Arena}},
  author={Zheng, Lianmin and Chiang, Wei-Lin and Sheng, Ying and Zhuang, Siyuan and Wu, Zhanghao and Zhuang, Yonghao and Lin, Zi and Li, Zhuohan and Li, Dacheng and Xing, Eric and others},
  booktitle={Advances in Neural Information Processing Systems},
  volume={36},
  pages={46595--46623},
  year={2023},
  url = {https://arxiv.org/abs/2306.05685}
}

@inproceedings{liu2025your,
  title={{Is Your Video Language Model a Reliable Judge?}},
  author={Liu, Ming and Zhang, Wensheng},
  booktitle={International Conference on Learning Representations},
  year={2025},
  url = {https://proceedings.iclr.cc/paper_files/paper/2025/hash/dc4f891373d19087d1ddda33e81e00e4-Abstract-Conference.html}
}

@article{hu2025timejudge,
  title={{TimeJudge: Empowering Video-LLMs as Zero-shot Judges for Temporal Consistency in Video Captions}},
  author={Hu, Yangliu and Song, Zikai and Yu, Junqing and Chen, Yiping Phoebe and Yang, Wei},
  journal={Frontiers of Information Technology \& Electronic Engineering},
  volume={26},
  number={11},
  pages={2204--2214},
  year={2025},
  url = {https://jzus.zju.edu.cn/iparticle.php?doi=10.1631%2FFITEE.2500412}
}

@inproceedings{tripathi2025pairwise,
  title={{Pairwise or Pointwise? Evaluating Feedback Protocols for Bias in LLM-Based Evaluation}},
  author={Tripathi, Tuhina and Wadhwa, Manya and Durrett, Greg and Niekum, Scott},
  booktitle={Conference on Language Modeling},
  year={2025},
  url = {https://arxiv.org/abs/2504.14716}
}

@inproceedings{li2024benchmarking,
  title={{Benchmarking and Improving Generator-Validator Consistency of Language Models}},
  author={Li, Xiang Lisa and Shrivastava, Vaishnavi and Li, Siyan and Hashimoto, Tatsunori and Liang, Percy},
  booktitle={International Conference on Learning Representations},
  year={2024},
  url = {https://arxiv.org/abs/2310.01846}
}

@inproceedings{liu2023g,
  title={{G-Eval: NLG Evaluation using GPT-4 with Better Human Alignment}},
  author={Liu, Yang and Iter, Dan and Xu, Yichong and Wang, Shuohang and Xu, Ruochen and Zhu, Chenguang},
  booktitle={Proceedings of the Conference on Empirical Methods in Natural Language Processing},
  pages={2511--2522},
  year={2023},
  url = {https://aclanthology.org/2023.emnlp-main.153/}
}

@inproceedings{bagad2023test,
  title={{Test of Time: Instilling Video-Language Models with a Sense of Time}},
  author={Bagad, Piyush and Tapaswi, Makarand and Snoek, Cees G. M.},
  booktitle={Proceedings of the IEEE/CVF Conference on Computer Vision and Pattern Recognition},
  pages={2503--2516},
  year={2023},
  url = {https://openaccess.thecvf.com/content/CVPR2023/html/Bagad_Test_of_Time_Instilling_Video-Language_Models_With_a_Sense_of_CVPR_2023_paper.html}
}

@article{zhang2024vinoground,
  title={{Vinoground: Scrutinizing LMMs over Dense Temporal Reasoning with Short Videos}},
  author={Zhang, Jianrui and Cai, Mu and Lee, Yong Jae},
  journal={arXiv preprint arXiv:2410.02763},
  year={2024},
  url = {https://arxiv.org/abs/2410.02763}
}

@inproceedings{sanyal2024machines,
  title = {{Are Machines Better at Complex Reasoning? Unveiling Human-Machine Inference Gaps in Entailment Verification}},
  author = {Soumya Sanyal and Tianyi Xiao and Jiacheng Liu and Wenya Wang and Xiang Ren},
  booktitle = {Findings of the Association for Computational Linguistics: ACL 2024},
  pages = {10361--10386},
  year = {2024},
  url = {https://aclanthology.org/2024.findings-acl.618/}
}

@inproceedings{saravanan2025velociti,
  title={{VELOCITI: Benchmarking Video-Language Compositional Reasoning with Strict Entailment}},
  author={Saravanan, Darshana and Gupta, Varun and Singh, Darshan and Khan, Zeeshan and Gandhi, Vineet and Tapaswi, Makarand},
  booktitle={Proceedings of the IEEE/CVF Conference on Computer Vision and Pattern Recognition},
  pages={18914--18924},
  year={2025},
  url = {https://openaccess.thecvf.com/content/CVPR2025/html/Saravanan_VELOCITI_Benchmarking_Video-Language_Compositional_Reasoning_with_Strict_Entailment_CVPR_2025_paper.html}
}

@inproceedings{bansal2024videocon,
  title={{VideoCon: Robust Video-Language Alignment via Contrast Captions}},
  author={Bansal, Hritik and Bitton, Yonatan and Szpektor, Idan and Chang, Kai-Wei and Grover, Aditya},
  booktitle={Proceedings of the IEEE/CVF Conference on Computer Vision and Pattern Recognition},
  pages={13927--13937},
  year={2024}
}

@inproceedings{li2024naturalbench,
  title={{NaturalBench: Evaluating Vision-Language Models on Natural Adversarial Samples}},
  author={Li, Baiqi and Lin, Zhiqiu and Peng, Wenxuan and Nyandwi, Jean De Dieu and Jiang, Daniel and Ma, Zixian and Khanuja, Simran and Krishna, Ranjay and Neubig, Graham and Ramanan, Deva},
  booktitle={Advances in Neural Information Processing Systems},
  volume={37},
  pages={17044--17068},
  year={2024}
}

@inproceedings{li2024vitatecs,
  title={{VITATECS: A Diagnostic Dataset for Temporal Concept Understanding of Video-Language Models}},
  author={Li, Shicheng and Li, Lei and Liu, Yi and Ren, Shuhuai and Liu, Yuanxin and Gao, Rundong and Sun, Xu and Hou, Lu},
  booktitle={European Conference on Computer Vision},
  pages={331--348},
  year={2024}
}

@article{li2026timeblind,
  title={{TimeBlind: A Spatio-Temporal Compositionality Benchmark for Video LLMs}},
  author={Li, Baiqi and Zhao, Kangyi and Zhang, Ce and Mitra, Chancharik and Nyandwi, Jean de Dieu and Bertasius, Gedas},
  journal={arXiv preprint arXiv:2602.00288},
  year={2026},
  url = {https://arxiv.org/abs/2602.00288}
}

@inproceedings{yuksekgonul2022and,
  title={{When and Why Vision-Language Models Behave like Bags-of-Words, and What to Do About It?}},
  author={Yuksekgonul, Mert and Bianchi, Federico and Kalluri, Pratyusha and Jurafsky, Dan and Zou, James},
  booktitle={International Conference on Learning Representations},
  year={2023},
  url = {https://arxiv.org/abs/2210.01936}
}

@article{han2026tracing,
  title={{Tracing the Arrow of Time: Diagnosing Temporal Information Flow in Video-LLMs}},
  author={Han, Peitao and Cheng, Fei and Pereira, Lis K and Liu, Qianying and Kitazawa, Shigeru},
  journal={arXiv preprint arXiv:2605.07568},
  year={2026},
  url = {https://arxiv.org/abs/2605.07568}
}

@article{xiao2025videoqa,
  title={{VideoQA in the Era of LLMs: An Empirical Study}},
  author={Xiao, Junbin and Huang, Nanxin and Qin, Hangyu and Li, Dongyang and Li, Yicong and Zhu, Fengbin and Tao, Zhulin and Yu, Jianxing and Lin, Liang and Chua, Tat-Seng and others},
  journal={International Journal of Computer Vision},
  volume={133},
  number={7},
  pages={3970--3993},
  year={2025},
  url = {https://link.springer.com/article/10.1007/s11263-025-02385-8}
}

@inproceedings{fu2025video,
  title={{Video-MME: The First-Ever Comprehensive Evaluation Benchmark of Multi-modal LLMs in Video Analysis}},
  author={Fu, Chaoyou and Dai, Yuhan and Luo, Yongdong and Li, Lei and Ren, Shuhuai and Zhang, Renrui and Wang, Zihan and Zhou, Chenyu and Shen, Yunhang and Zhang, Mengdan and others},
  booktitle={Proceedings of the IEEE/CVF Conference on Computer Vision and Pattern Recognition},
  pages={24108--24118},
  year={2025},
  url = {https://openaccess.thecvf.com/content/CVPR2025/html/Fu_Video-MME_The_First-Ever_Comprehensive_Evaluation_Benchmark_of_Multi-modal_LLMs_in_CVPR_2025_paper.html}
}

@article{alain2016probes,
  title={{Understanding Intermediate Layers Using Linear Classifier Probes}},
  author={Alain, Guillaume and Bengio, Yoshua},
  journal={arXiv preprint arXiv:1610.01644},
  year={2016},
  url = {https://arxiv.org/abs/1610.01644}
}

@inproceedings{burns2023latent,
  title={{Discovering Latent Knowledge in Language Models Without Supervision}},
  author={Burns, Collin and Ye, Haotian and Klein, Dan and Steinhardt, Jacob},
  booktitle={International Conference on Learning Representations},
  year={2023},
  url = {https://arxiv.org/abs/2212.03827}
}

@inproceedings{liu2026seeing,
  title={{Seeing but Not Believing: Probing the Disconnect Between Visual Attention and Answer Correctness in VLMs}},
  author={Liu, Zhining and Chen, Ziyi and Liu, Hui and Luo, Chen and Tang, Xianfeng and Wang, Suhang and Zeng, Jingying and Dai, Zhenwei and Shi, Zhan and Wei, Tianxin and others},
  booktitle={International Conference on Learning Representations},
  year={2026},
  url = {https://arxiv.org/abs/2510.17771}
}

@inproceedings{leng2024mitigating,
  title={{Mitigating Object Hallucinations in Large Vision-Language Models through Visual Contrastive Decoding}},
  author={Leng, Sicong and Zhang, Hang and Chen, Guanzheng and Li, Xin and Lu, Shijian and Miao, Chunyan and Bing, Lidong},
  booktitle={Proceedings of the IEEE/CVF Conference on Computer Vision and Pattern Recognition},
  pages={13872--13882},
  year={2024},
  url = {https://openaccess.thecvf.com/content/CVPR2024/html/Leng_Mitigating_Object_Hallucinations_in_Large_Vision-Language_Models_through_Visual_Contrastive_CVPR_2024_paper.html}
}

@inproceedings{wang2026calibrating,
  title={{Calibrating Verbalized Confidence with Self-Generated Distractors}},
  author={Wang, Victor and Stengel-Eskin, Elias},
  booktitle={International Conference on Learning Representations},
  year={2026},
  url = {https://arxiv.org/abs/2509.25532}
}

@misc{openai2025gpt41,
  title={{Introducing GPT-4.1 in the API}},
  author={{OpenAI}},
  howpublished={\url{https://openai.com/index/gpt-4-1/}},
  year={2025}
}

@misc{anthropic2025claude,
  title={{Claude}},
  author={{Anthropic}},
  howpublished={\url{https://www.anthropic.com/claude}},
  year={n.d.},
  note={Accessed November 7, 2026}
}
